\documentclass[letterpaper,journal]{IEEEtran}
\usepackage{amsmath,amsfonts}
\usepackage{algorithmic}
\usepackage{algorithm}
\usepackage{enumitem}
\usepackage{array}
\usepackage[caption=false,font=footnotesize]{subfig}
\usepackage{textcomp}
\usepackage{stfloats}
\usepackage{xcolor}
\usepackage{hyperref}
\usepackage{url}
\usepackage{verbatim}
\usepackage{graphicx}
\usepackage{booktabs}
\usepackage{epstopdf}
\usepackage{cite}

\hypersetup{
  colorlinks=true, % 将链接设置为彩色
  linkcolor=red, % 设置链接颜色为红色
  citecolor=green, % 设置引用颜色为红色
  urlcolor=red % 设置 URL 颜色为蓝色
}
\begin{document}

\title{Zero-Shot Enhancement of Low-Light Image \\Based on Retinex Decomposition}
%\author{Wenchao Li,Bangshu Xiong,Xiaoyun Long,jiabao Chen, ,~\IEEEmembership{Staff,~IEEE,}
\author{Wenchao Li, Bangshu Xiong, Qiaofeng Ou, Xiaoyun Long, Jinhao Zhu, Jiabao Chen,and Shuyuan Wen
        % <-this % stops a space
\thanks{This study was partially supported by the National Natural Science Foundation of China under Grant Nos. 61866027, 62365014, and 62162044, the Key Research and Development Program of Jiangxi Province under Grant No. 20212BBE53017, and the Aviation Science Foundation under Grant No. 20200057056006.}% <-this % stops a space
\thanks{Wenchao Li, Jinhao Zhu, and Shuyuan Wen are with Key Laboratory of Precision Opto-mechatronics Technology, Ministry of Education, and the School of Instrumentation Science and Opto-electronics Engineering, Beihang University, Beijing 100191, People’s Republic of China, (e-mail: wenchaoli@buaa.edu.cn; jinhaozhu@buaa.edu.cn; shuyuanwen@buaa.edu.cn).

 Bangshu Xiong , Qiaofeng Ou, and Jiabao Chen are with the School of Information Engineering, Nanchang Hangkong University, Nanchang 330063, People’s Republic of China (e-mail: xiongbs@126.com; ou.qiaofeng@nchu.edu.cn; jiabaoc2022@163.com).
 
 Xiaoyun Long is with Department of Electrical Engineering, Ganzhou Vocational and Technical College, Ganzhou 341000, People’s Republic of China,(e-mail:l2433947465@163.com).}% <-this % stops a space
\thanks{Manuscript received April 19, 2021; revised August 16, 2021.}}

\markboth{Journal of \LaTeX\ Class Files,~Vol.~14, No.~8, August~2021}%
{Shell \MakeLowercase{\textit{et al.}}:Zero-shot enhancement of low-light image \\based on retinex decomposition}


\maketitle

\begin{abstract}
There are two difficulties that make microlight image enhancement a challenging task; firstly, it needs to consider not only luminance restoration but also image contrast, image denoising and color distortion issues simultaneously. Second, the effectiveness of existing low-light enhancement methods depends on paired or unpaired training data with poor generalization performance. To solve these difficult problems, we propose in this paper a new learning-based Retinex decomposition of zero-shot low-light enhancement method, called ZERRINNet. To this end, we first designed the N-Net network, together with the noise loss term, to be used for denoising the original low-light image by estimating the noise of the low-light image. Moreover, RI-Net is used to estimate the reflection component and illumination component, and in order to solve the color distortion and contrast, we use the texture loss term and segmented smoothing loss to constrain the reflection component and illumination component. Finally, our method is a zero-reference enhancement method that is not affected by the training data of paired and unpaired datasets, so our generalization performance is greatly improved, and in the paper, we have effectively validated it with a homemade real-life low-light dataset and additionally with advanced vision tasks, such as face detection, target recognition, and instance segmentation. We conducted comparative experiments on a large number of public datasets and the results show that the performance of our method is competitive compared to the current state-of-the-art methods. The code is available at:\href{https://github.com/liwenchao0615/ZERRINNet}
{https://github.com/liwenchao0615/ZERRINNet}
\end{abstract}


\section{INTRODUCTION}
\IEEEPARstart{D}{ue} to the lack of ambient light, images taken are often low-light images. These images are characterized by low contrast, high noise, and unclear details. These issues not only impact human visual perception but also hinder the application of advanced computer vision tasks, such as autonomous driving and human behavior recognition. Thus, improving the quality of these shimmering images is necessary to increase the accuracy of advanced visual tasks in low-light scenes.

Improving the quality of low-light images involves addressing two challenges: enhancing image brightness and restoring image details. However, boosting image brightness often introduces issues such as noise, color distortion, and region blurring, ultimately resulting in the loss of original detail information. In recent years, researchers have proposed a series of effective low-light enhancement methods, such as histogram equalization, Retinex theory and deep learning enhancement techniques \cite{zheng2022low}, \cite{2021survry}. However, none of these methods can effectively address both major problems at the same time. Among the various methods used to enhance low-light images, the most convincing one is based on the Retinex decomposition theory and its improved versions. According to the Retinex theory, an image can be divided into two parts: a light component and a reflection component. The reflection component is an inherent characteristic of the image and remains unchanged regardless of variations in the external ambient illumination. In contrast, the illumination component varies with changes in the external environment.

Proposed are several low-light image enhancement methods based on the principle of reflection component invariance of Retinex. For instance, the direct use of the reflection component as the enhancement result is a simple and effective method for low-light images. However, this approach results in a loss of some image detail information. To address this issue, one can brighten the illumination component obtained through decomposing the low-light image and then multiply it with the reflection component to obtain the final enhancement result. This method effectively mitigates the loss of details that occur when the reflection component is directly used as the enhancement result. However, none of these methods take into account the noise of the image \cite{land1971lightness}, \cite{MSR}, \cite{MSRCR}, \cite{zhang2021better}.

Considering that traditional Retinex methods require more parameters and denoising of images, people combine deep learning and Retinex theory. Chen et al. \cite{retinex-net}  proposed a convolutional neural network model based on Retinex theory and introduced the traditional BM3D method in the adjustment module to remove noise, which unfortunately produced problems, such as color distortion in the enhancement process. Recently, Wu et al. \cite{uretinex}  formulated the decomposition problem as an implicit a priori regularization model. This approach effectively performs noise suppression and detail preservation. However, these methods require the preparation of paired datasets, which limits the further use of the methods.

\begin{figure}[!tbp]
\centering
\includegraphics[width=\linewidth]{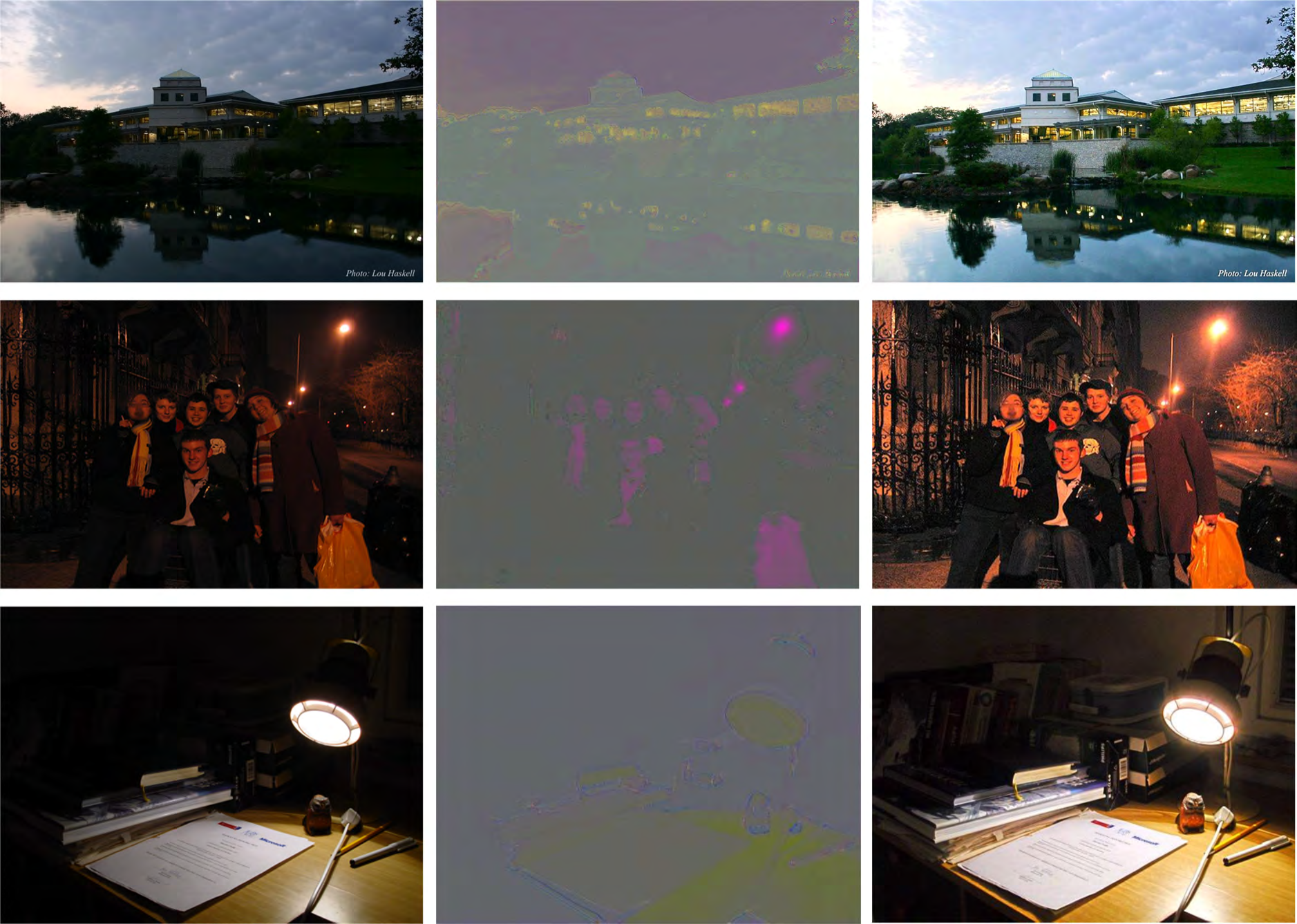}
\caption{ Restoration results of ZERRINNet. Images in the left column are underexposed images in different scenes. In the center is the extracted noise image, while the ones in the right column are the restored results.}
\label{one:image}
\end{figure}

Existing low-light enhancement methods do not simultaneously address the problems of image noise, color distortion, and the results are heavily dependent on the training of paired and unpaired datasets. To address these problems, this paper proposes a simple zero-shot unified learning-based Retinex low-light enhancement framework. The method is based on Eq.\ref{one:eq} dividing the network structure into three modules: input, Retinex decomposition and image noise extraction and the final enhancement part.
\begin{equation}
{E}(x)={R}(x) \odot {I}(x)+{N}(x),
\label{one:eq}
\end{equation}
specifically, for the Retinex decomposition and noise extraction modules, the process extracts the reflection component ${R}(x)$, the illumination component ${I}(x)$ and the noise component ${N}(x)$ , respectively. The input images are not identical in both, where the noise component ${N}(x)$  is extracted from the original shimmering image. This is because denoising directly in the reflection component will inevitably lead to the loss of detail information. At the same time, we fused the original shimmer image with the V-channel image as an input to the Retinex decomposition to increase the detail information of the image. In order to solve the problem of color distortion in the image, we design a texture enhancement term, and this loss term can effectively guide the image decomposition, as shown in Fig \ref{one:image}. 

Retinex decomposition removes primarily the illumination component and enhances the image using the reflection component. Alternatively, enhancement results are obtained by adjusting the illumination component and multiplying it by the reflection component. However, Retinex decomposition tends to generate a lot of noise, which is a difficult problem. One has mainly addressed this problem by denoising on the reflection component and enhancing the contrast of the illumination component. However, there are still some unresolved issues. First, it cannot handle color distortion, denoising and image contrast simultaneously. Second, the method requires paired datasets as training data, limiting its application to advanced vision tasks.

The main contributions of this paper can be summarized as follows:
1) We propose an elegant unified zero-reference depth framework based on the combination of retinex decomposition and deep learning, called ZERRINNet, for shimmering image enhancement that avoids the challenge of collecting paired datasets.
2) Taking into account the noise and color distortion problems in Retinex decomposition, a noise extraction network N-Net is designed to remove noise from shimmering images, and a loss term of texture enhancement is designed to guide Retinex decomposition.
3) Comprehensive experiments were conducted, and the experimental results show that our method largely outperforms the state-of-the-art methods. In addition, we test our method on a homemade dataset of real shimmering images, face detection, target recognition, and instance segmentation, demonstrating the robustness and generalization ability of our method.

The rest of the paper is organized as follows. Section II briefly reviews related work on the low image enhancement problem. Section III briefly describes the network structure and loss function of the proposed method, followed by specific experimental details in Section IV. We quantitatively and qualitatively evaluate the performance of the proposed method and test the generalization performance of the proposed method in Section V. Section VI summarizes the paper.

\section{RELATED WORK}
 In this section, we briefly review the development of low-light image enhancement techniques from the very beginning of non-learning-based image enhancement methods such as histogram equalization and Retinex methods, then, current learning-based image enhancement methods, finally, we also introduce low-light image enhancement methods that combine Retinex and deep learning.

\textbf{Non-learning based image enhancement methods.} Non-learning based image enhancement methods, currently, mainly consist of histogram equalization or Retinex models and improvements of the above methods. Histogram Equalization (HE) enhancement method aims to adjust the pixel values of an image to change the contrast and brightness distribution of the image. however, using this method is to adjust the whole image, which inevitably leads to the details of the image being over-enhanced or compressed, which makes the whole image look unnatural or distorted, in addition, it is not a good way to enhance the image. since the histogram equalization is adjusted based on the frequency of the pixel values, it amplifies the noise to some extent. To address these problems, corresponding prior and constraints have been proposed to be attached to the histogram equalization. Such as local histogram equalization methods \cite{reza2004realization}, noise suppression methods \cite{hedenoiseimage}, brightness adjustment methods \cite{hebrightness}, dark detail enhancement methods \cite{dackhe} and computational efficiency improvement methods \cite{pizer1987adaptive}. Although the above methods have achieved certain results, there are still some problems for the detail recovery of the image, there are over-enhancement phenomena, and the direct application of low-light image enhancement still needs to be improved.

Another strategy is the low-light image enhancement method based on the Retinex model, which can effectively protect the detail information of the image and improve the contrast of the image, which is a model that simulates the mechanism of the human eye's perception of light and splits the low-light image into a reflection component and a light component. The most typical methods are Single Scale Retinex (SSR) and Multi Scale Retinex (MSR)\cite{MSR}, and based on these works, some variants have been successively proposed. Wang et al. \cite{wang2013} used a high pass filter to decompose the image to estimate the illumination map. LIME \cite{LIME} used a structure-aware smoothing model to estimate the illumination map after backprojection to obtain the enhancement results. Mading et al. \cite{Mading} estimated the noise distribution by modeling the noise distribution,  then estimated the reflectance and illumination components to effectively enhance the image. Low-light images accompanied by strong noise. However, the above Retinex methods, despite having interpretable physical models, these methods need to rely on hand-designed prior and regularization terms, and it is worth noting that finding a valid prior or regularization is challenging and inaccurate prior or regularization may lead to artifacts and color bias in the enhancement results,  furthermore, it can be found that most of the Retinex-based methods do not take noise into account or even amplify it.

\textbf{Learning-based enhancement methods.}
In recent years, deep learning has achieved impressive success in the field of low-light image enhancement with fewer parameters, robustness and speed compared to nonlearning low-light image enhancement methods~\cite{li2021low},~\cite{ren2018lecarm}. This has led to an increased interest in learning-based low-light image enhancement, and to the best of our knowledge, LLNet \cite{LLNet} is the first approach to solve the low-light image enhancement problem using a deep neural network, which employs a variant of the superimposed sparse denoising auto-encoder for both brightening and denoising of low-light images. MBLLEN \cite{MBLLEN} designed an end-to-end feature extraction, enhancement, and fusion in three modules to enhance the image, and subsequent authors \cite{attention} further improved the method by proposing three subnetworks for detail enhancement, denoising and reconstruction of dark regions, respectively. Xu et al. \cite{xu2020learning} proposed a frequency-based decomposition and enhancement network. This network is able to recover the content and details of the image while suppressing the noise.
UTVNet \cite{utvnet} obtained the final enhancement result by constructing a noise model for low-light images and then brightening and denoising the images. LLFIow \cite{llFlow} designed a novel stream regularization model to solve the low-light image enhancement problem. The above low-light image enhancement methods, despite some progress in image enhancement, unfortunately require the preparation of pairs of datasets, which are pairs of datasets, which are more difficult to photograph and rely on, which also leads to a weakening of the model's generalization ability. Therefore, semi-supervised and unsupervised enhancement methods have been proposed accordingly.

DRBN \cite{DRBN} is the first to introduce semi-supervised learning to low-light image enhancement, which can enhance the visual quality of images under unpaired datasets, however, it still needs to resort to paired datasets to learn the details and signal fidelity under individual frequency band signals. To avoid this problem, the EnlightenGAN \cite{enlightengan} method is proposed, which utilizes the illuminance information of low-light inputs as a self-regularized attentional map at each layer of the depth feature to regulate the unsupervised learning. It is an efficient unsupervised generative adversarial network. SCI \cite{sci} constructs a weight-sharing illumination learning process with self-calibrating modules, which eliminates the cumbersome process of designing the network structure and achieves the purpose of enhancement using only simple operations. These methods have achieved very competitive performance in image enhancement, but still face some challenges, such as how to achieve stable training, how to avoid color bias, and how to establish connectivity relationships across domain information.

\textbf{Deep Retinex based methods.} In recent times, considering the respective advantages of Retinex theory and learning-based neural networks, people have developed some better enhancement methods by combining the two effectively \cite{yu2017low},\cite{xu2022structure}, \cite{rrdnet},~\cite{liang2022self}. LightenNet \cite{LightenNet}  designed a lightweight network model to take low-light images as input, go through image preprocessing, network training, post-processing, and finally get the enhancement results. In order to accurately decompose the reflection component and the off lighting component, Zhang et al.\cite{kind} designed the KinD network, which is divided into three modules responsible for the original image decomposition, reflectivity recovery, and lighting adjustment, and finally gets the enhancement result. The authors added a multiscale illumination attention module \cite{kind++} in order to compensate for visual defects caused by post-amplification of the H-dark region. Wang et al. \cite{wang} proposed a progressive Retinex network framework, where one network is responsible for estimating the illumination level and the other for estimating the noise level. The above two sub-networks work in an asymptotic manner until stable results are obtained.
Fan et al. \cite{fan_segmentation} combined semantic segmentation and Retinex modeling to guide the enhancement of illumination and reflection components through a priori information about image semantics. Considering that data-driven approaches may provide undesired enhancement results, URetinex-Net \cite{uretinex} cleverly transforms the optimization problem into a learnable network by adaptively fitting an implicit prior in a data-driven manner, achieving noise suppression and detail preservation in the final decomposition result, and also achieving a more accurate and efficient decomposition. Noise suppression and detail preservation, and also allows flexible illumination enhancement through user customization. Surely, deep Retinex-based methods can achieve better enhancement performance in general, however, all of them need to face the respective sufferings of paired and unpaired data, and recently some zero-shot learning \cite{ExCNet},\cite{guo2020zero},\cite{retinexdip},\cite{Zero-DCE++},\cite{RUAS},\cite{SGZ}, methods have been mentioned, which learns enhancement only from test images,  obviously, it does not require paired or unpaired training data, and thus can achieve excellent generalization ability that can be better applied in various scenarios.

\section{PPROPOSED APPROACH}

\subsection{Network Architecture}
\begin{figure*}[!t]
  \centering
  \includegraphics[width=\textwidth]{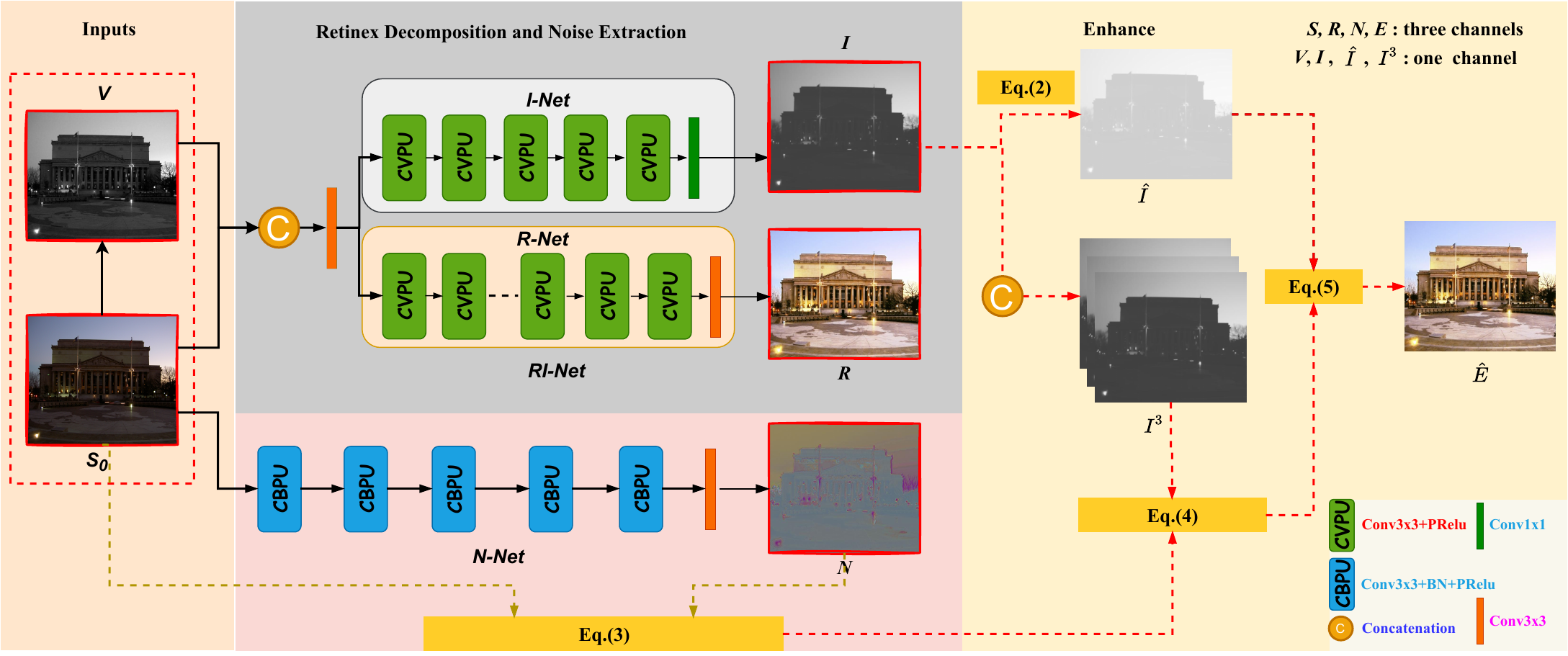}
  \caption{Overall unsupervised low-light image enhancement based on Retinex decomposition and denoising framework of ZERRINNet. $I_1$, R and N represent the illumination reflectance and noise, respectively. S is the input low-light image, $V$ is v-channel image of the original image ,  $\hat{E}$ is Enhancement of images.}
  \label{fig:two}
\end{figure*}

In the field of low-light enhancement, enhancement methods based on the combination of Retinex theory and deep learning have been widely used. However, for these enhancement methods, on the one hand, it has been found that some methods do not take into account the noise problem of the image during the image recovery process, and even if denoising methods are added subsequently, the results are not satisfactory, and even lead to artifacts and color distortion in the reconstructed image, and do not effectively deal with the problems of contrast, denoising and color distortion at the same time. On the other hand, some enhancement methods rely on paired and unpaired dataset training with poor generalization performance, which is not conducive to the generalization of the methods. For the above motivation, in this paper, a unified Retinex-based zero-shot framework for deep learning, namely ZERRINNet, is proposed. The overall framework of ZERRINNet is shown in Fig \ref{fig:two}. As stated earlier, we divide the whole framework into three parts, input, image decomposition, noise extraction, and final enhancement based on Eq \ref{one:eq}. Firstly, we carefully designed two sub-networks, RI-Net and N-Net, which are used for Retinex decomposition and noise extraction of the input image, however, we find that the reflection component contains more information than the illumination component, and in order to fully learn the features of the reflection component from the input image,  we deepened the R-Net network used for extracting the reflection, compared to the I-Net network used for extracting the illumination component. We deepened the R-Net network used to extract the reflection compared to the I-Net network used to extract the illumination component. Moreover, to enhance the detail information and avoid artifacts, on one hand, the inputs of the networks are slightly different, we first extract the V-channel of the low-light image, then fuse the V-channel with the original shimmering image, and then use the fused synthesized image as an input to the RI-Net network. In the shimmer image noise extraction, we take the original shimmer image itself as the input to the network. In addition, in order to extract as much integrated noise as possible from the original shimmering image, we compare some existing denoising methods in our experiments, and we use multilayer convolution plus BN and activation structure.

In contrast to existing deep learning based image enhancement methods \cite{hai2023r2rnet},\cite{kind++}, our denoising strategy denoises the original low-light image instead of the reflective component with the aim of avoiding loss of image information, furthermore, our method does not need to be trained on any external image but tests the input image directly to obtain the enhanced image. Our method eliminates the a priori knowledge learned from training on external datasets and enables the ZERRINNet network to realize the mapping from low light images to enhanced images by means of a specially designed loss function.

\subsection{Loss Functions}
We need to not only improve the contrast degree but also take into account image noise and color distortion, based on our decomposition strategy and enhancement approach, the loss function used in this paper is simple and common as described below.
We designed an overall loss function to evaluate the decomposition and denoising of the network to guide the network to generate more accurate components. We design a loss function $\mathcal{L}_{\mathrm{total loss}}$ that consists of three parts as,
\begin{equation}
\label{eq:5}
\mathcal{L}_{\mathrm{total loss}}=\mathcal{L}_{\mathrm{recon}}+\mathcal{L}_{\mathrm{smooth}}+
\mathcal{L}_{\mathrm{texture}}+\mathcal{L}_{\mathrm{noise}},
\end{equation}
where $\mathcal{L}_{recon}$, $\mathcal{L}_{\mathrm{smooth}}$,
$\mathcal{L}_{\mathrm{texture}}$ and $\mathcal{L}_{\mathrm{noise}}$ corresponding to the reconstruction loss, smoothing loss,  image noise loss and  texture enhancement loss.

\textbf{Reconstruction loss.} The decomposition strategy of the image needs to meet the requirements of reconstructing the image, so that a reasonable decomposition result can be obtained, the reconstruction loss design is based on the basic idea of Retinex decomposition, which not only needs to consider the two ${R}(x_1)$ and ${I}(x_1)$ components, but also needs to add the noise of the original low-light image. Therefore, in order to ensure the similarity between the observed image and the reconstructed image, The reconstruction loss is defined as,
\begin{equation}
\mathcal{L}_{\text {recon}}=|({R}(x_1) \odot {I}(x_1)+{N}(x_0))-{S}_0|_1,
\end{equation}
where $x_1$ and $x_0$ indicate the Synthesized image of low-light image fused with V-channel and original low-light image.

\textbf{Smoothing loss.} For natural images, illumination is generally locally smooth. On the contrary, for low-light images, the image content and details tend to be weak and the contrast is low. Smoothness loss encourages illumination to be smooth on pixels with small gradients and discontinuous on pixels with large gradients. Therefore, we use segmented smoothness loss to estimate the illumination component and reflection component to improve the contrast of the image. The total loss function can be defined as,
\begin{equation}
\mathcal{L}_{smooth} ={\lambda_i}\mathcal{L}_{is} + {\lambda_r}\mathcal{L}_{rs} ,
\end{equation}
where $\lambda_i$ and $\lambda_r$, corresponding to the coefficients of both segment-smoothed light loss and segment-smoothed reflection loss, respectively. $\mathcal{L}_{is}$ and $\mathcal{L}_{rs}$ indicated are segment-smoothed light loss and segment-smoothed reflection loss.
Segmented smoothed light mapping recovers better detail information in dark regions, this is because when neighboring pixels have similar intensity, their contrast is amplified at the same illumination value. In addition, the L1 loss is used to calculate the difference between the illumination component and the maximum value of the original low-light image. The overall smoothing loss of the illumination component can be expressed as
\begin{equation}
\mathcal{L}_{is} = \sum_{h,w}{w}(x_1) | {\nabla}^{(h,w)}{I}(x) |+  | {I}(x)-{S}_m |_1 ,
\end{equation}
where  ${S}_m$ is the maximum channel of the original low-light image , here $\nabla^{(h,w)}$ is the gradient in the horizontal and vertical directions, ${w}(x_1)$ is the weight function . According to the RTV loss \cite{xu2012structure} , it is shown that the weight term is inversely proportional to the gradient. Therefore, the weight function we designed can be expressed as
\begin{equation}
{w}(\mathbf{x_1})=\frac{1}{G {\circ}(\partial I_g(\mathbf{x_1}))^2},
\end{equation}
where ${G}$ is a Gaussian filter, $\circ$ denotes the convolution operator and ${I_g}$ is the gray-scale version of the input. $\partial$ is the parameters.

Similarly, segmented reflection smoothing loss is similar to illumination loss in that it aims to recover structural and detail information, and unlike illumination smoothing, it focuses on points with small horizontal and vertical gradients, ensuring that the true noise points are smoothed rather than the edges, which nonetheless need to be constrained by the illumination map. The overall equation can be expressed as
\begin{align}
\mathcal{L}_{rs} &= \sum_{h,w}{w}_r(x_1) {\nabla}^{(h,w)}{R}(x_1) | \nonumber \ \\ &\quad + \lambda_{rs}|({S_0}/{I}(x_1))-{R}(x_1) |_1 ,
\end{align}
where $\lambda_{rs}$ is the coefficient of loss of this reflected component, ${w}_r(x_1)$ is the weight parameter of the gradient of the reflection component, it uses normalize to denote min-max normalization with the following equation
\begin{equation}
\boldsymbol{w}_r(\mathbf{x})=\operatorname{normalize}\left(\frac{1}{{I}(x_1) \cdot \nabla^{(h,w)}{S_0} + \epsilon }\right),
\end{equation}
where $\nabla^{(h,w)}{S_0}$ is Horizontal and vertical gradients in raw low light, $\epsilon$ then as a constant, its value is 0.0001.

\textbf{Texture enhancement loss.}
In the process of low-light image reconstruction, the problems of color distortion and blurring of details in dark areas are often faced. Therefore, in this paper, a texture enhancement loss is designed which consists mainly of color loss and dark area loss \cite{SGZ}. Color loss \cite{wang2019underexposed} reduces the color errors in the enhanced image by bridging different color channels. We use Charbonnier loss which helps in high quality image reconstruction, extract region loss to improve the quality of the shimmering region considering the uneven illumination of the shimmering image, and segment the darkest pixel value of the image by 40\% (empirical value) to approximate the dark region of the whole image, and also limit the value of the reflectance map to the range [0, 1] to avoid over-enhancement. The total texture enhancement loss can be expressed as,
\begin{equation}
\mathcal{L}_{texture} =\mathcal{L}_{color} + \mathcal{L}_{region}+\mathcal{L}_{maxa} ,
\end{equation}
where $\mathcal{L}_{color}$, $\mathcal{L}_{region}$ and $\mathcal{L}_{maxa}$ represent color loss , low-light area loss and low-light attention loss.

The color loss is designed to ensure that color distortions are avoided in the enhanced results. It is defined as:
\begin{equation}
\begin{array}{r}
\mathcal{L}_{color}=\sum\limits_{\forall(i, j) \in \zeta} \sqrt{\left(\left(Y^i\right)-\left(Y^j\right)\right)^2+\varepsilon^2}, \\
\zeta=\{(R, G),(R, B),(G, B)\},
\end{array}
\end{equation}
where $\epsilon$ is a penalty term that is empirically set to $10^{-6}$ for training stability.

The region loss is introduced to improve the quality of low-light areas, we use the way to segment images to build region loss, the region loss is defined as, 
\begin{align} 
\mathcal{L}_{region} &= w_L \cdot \frac{1}{m_L n_L} \sum_{i=1}^{n_L} \sum_{j=1}^{m_L}\left(\left|R_L(i, j)-{S_0}_L(i, j)\right|\right) \nonumber \ \\ &\quad + w_H \cdot \frac{1}{m_H n_H} \sum_{i=1}^{n_H} \sum_{j=1}^{m_H}\left(\left|R_H(i, j)-{S_0}_H(i, j)\right|\right), \end{align}
where $R_L$ and ${S_0}$ are the low-light regions of the reflection image and original low light image,  $R_H$ and ${S_0}_H$ are the rest parts of the images. In our case, we suggest $w_L$ = 4 and $w_H$ = 1.

In order to avoid over-enhancement in image recovery, we have added an attention loss to the original area loss, which limits the value of the reflectance map, allowing correct enhancement of underexposed areas and avoiding over-enhancement of normally exposed areas, the attention loss is defined as
\begin{equation}
\mathcal{L}_{maxa} =\frac{\left|\max _c(S_0)-\max _c({R}(x_1))\right|}{\max _c(S_0)},
\end{equation}
where $\max_{c}(x)$ returns the maximum value among three color channels, ${S_0}$ is the original bright image and ${R}(x_1)$ is the reflected image.

\textbf{Image noise loss.} During low-light image recovery, stretching the contrast enhances the details in the dark areas of the image,  at the same time, the noise hidden in the dark areas is amplified. Therefore, it is necessary to suppress the noise, especially the noise in the dark region. In this paper, we mainly design the loss function based on the designed noise extraction network, we use the illumination map of the image as the weights to extract the low-light image noise. The overall denoising loss function of the image is defined as
\begin{equation}
\mathcal{L}_{noise}=\left\|\boldsymbol{I}(x_1) \cdot {N}\right\|_F ,
\end{equation}
where ‖X‖F means the Frobenius norm of the matrix X, ${N}$ denotes the extracted noise . 
\subsection{Low-Light Image Enhancement}
The decomposition process of RI-Net and the noise extraction of the original shimmering image by N-Net were described in detail above. In this section, we describe how to perform the final enhancement of the image.
In this enhancement module, we are based on Eq \ref{one:eq} to design our enhancement idea. We refer to the Retinex decomposition enhancement strategy proposed by \cite{retinexdip}, and through experiments, we found that combining the adjusted illumination component with the denoised reflection component to get the final enhancement result has better results, for our method, Firstly, considering the enhancement of the contrast and brightness of the image, we convert the single-channel lightmap to a three-channel image, and then adjust the brightness of the three-channel lightmap by gamma transformation,
\begin{equation}
\widehat{I}(\mathbf{x})={I}(\mathbf{x})^\gamma ,
\label{two:eq}
\end{equation}
where $\gamma$ is a predefined parameter, $I(X)$ is three-channel lighting components, $\widehat{I}(\mathbf{x})$ is gamma-transformed lighting components, the noise of the low-light image is hidden in the reflection component, and it is necessary to denoise the reflection component. However, directly denoising the reflection component obtained by Retinex decomposition will cause the loss of image details. Therefore, according to the above description, this paper proposes denoising in the original low-light image, the noise-free raw low-light image can be computed as
\begin{equation} \widehat{S}(\mathbf{x})=S_0(\mathbf{x}) - N(\mathbf{x}) , \end{equation}
where $S_0(\mathbf{x})$ is the original low-light image, $N(\mathbf{x})$ and $\widehat{S}  (\mathbf{x})$ indicate the Noisy image and denoised low-light image. To suppress the amplification of noise, the reflection component $\widehat{R}(\mathbf{x})$ is obtained by dividing the original low light image without noise and the three-channel illumination component $I^3(\mathbf{x})$,
\begin{equation}
\label{eq:3}
\widehat{R}(\mathbf{x})=\widehat{S}(\mathbf{x}) / {I}^3(\mathbf{x}) , \end{equation}
the final enhancement result $\widehat{E}(\mathbf{x})$ is obtained by combining the illumination component of the three channels adjusted by the gamma transformation with the reflection component obtained, with the equation as follows: 
\begin{equation} \widehat{E}(\mathbf{x})=\widehat{R}(\mathbf{x}) \odot \widehat{I}(\mathbf{x}), 
\label{eq:15}
\end{equation}
in summary, Algorithm \ref{alg:alg1} describes the implementation of ZERRINNet. In a word, the algorithm first decomposes the input image into illumination and reflection components, while extracting the noise component of the original image, and then generates an enhanced image by locally adjusting the illumination component and then combining the adjusted illumination component with the denoised reflection component.

\begin{algorithm}[!tbp] 
\caption{ZERRINNet for low-light enhancement.}
\label{alg:alg1}
\renewcommand{\algorithmicrequire}{\textbf{Input:}}
\renewcommand{\algorithmicensure}{\textbf{Output:}}
\begin{algorithmic}[1]
    \REQUIRE Low-light image ${{S}_0}$, Low-light synthetic image ${S}$,  parameters  $\mathrm{\delta}$, $\mathrm{\gamma}$,  ${\mathrm{\lambda}_i}$, ${\mathrm{\lambda}_k}$, $\mathrm{\lambda}_n$, maximum iterations $\mathrm{M}$ %%input
    \ENSURE $\hat{E}$    %%output
\STATE \textbf{Step 1:} Synthetic low-light image Retinex decomposition  Original low-light noise extraction
\STATE \textbf{for} $m \gets 0$ \textbf{to} $M$ \textbf{do}
\STATE \hspace{0.5cm}$ N_k \gets {N}^k(x_2) ,x \in \{1,2\}$ \\
\hspace{0.5cm}$ I_k \gets {I}^k(x_1) , R_k \gets {R}^k(x_1) $
\STATE \hspace{0.5cm}Minimize the objective function Eq.~\textcolor{red}{\ref{eq:5}}
\STATE \hspace{0.5cm}Update ${N}^k$,${I}^k$ and ${R}^k$ simultaneously
\STATE \textbf{return} $\mathbf{N_k}$, $\mathbf{I_k}$ and $\mathbf{R_k}$
\STATE $ \mathbf{I}^3 \gets \text{Cat} \hspace{0.1cm} \mathbf{I}_k $
\STATE $ \hat{\mathbf{I}} \gets  \mathbf{I}_k^\gamma $
\STATE Produce processed reflection maps $\hat{\mathbf{R}^c}$  by denoising them with Eq.~\textcolor{red}{\ref{eq:3}}
\STATE Compute enhanced image $\hat{{E}}$ using Eq.~\textcolor{red}{\ref{eq:15}}
\STATE\textbf{return}\hspace{0.2cm} $\hat{{E}}$ 
\end{algorithmic}
\end{algorithm}

\section{EXPERIMENTS AND DISCUSSION}
In this section, we first visualize the entire flow of the proposed method to better understand our approach, and then describe the specific details of the experiment, including the public dataset and evaluation metrics used in the experiment.

\subsection{Study Our Algorithms}
\begin{figure}[!tbp]
    \centering
    \small
    \tabcolsep=0.05cm
    \begin{tabular}{c c c }
        \includegraphics[width=0.32\linewidth]{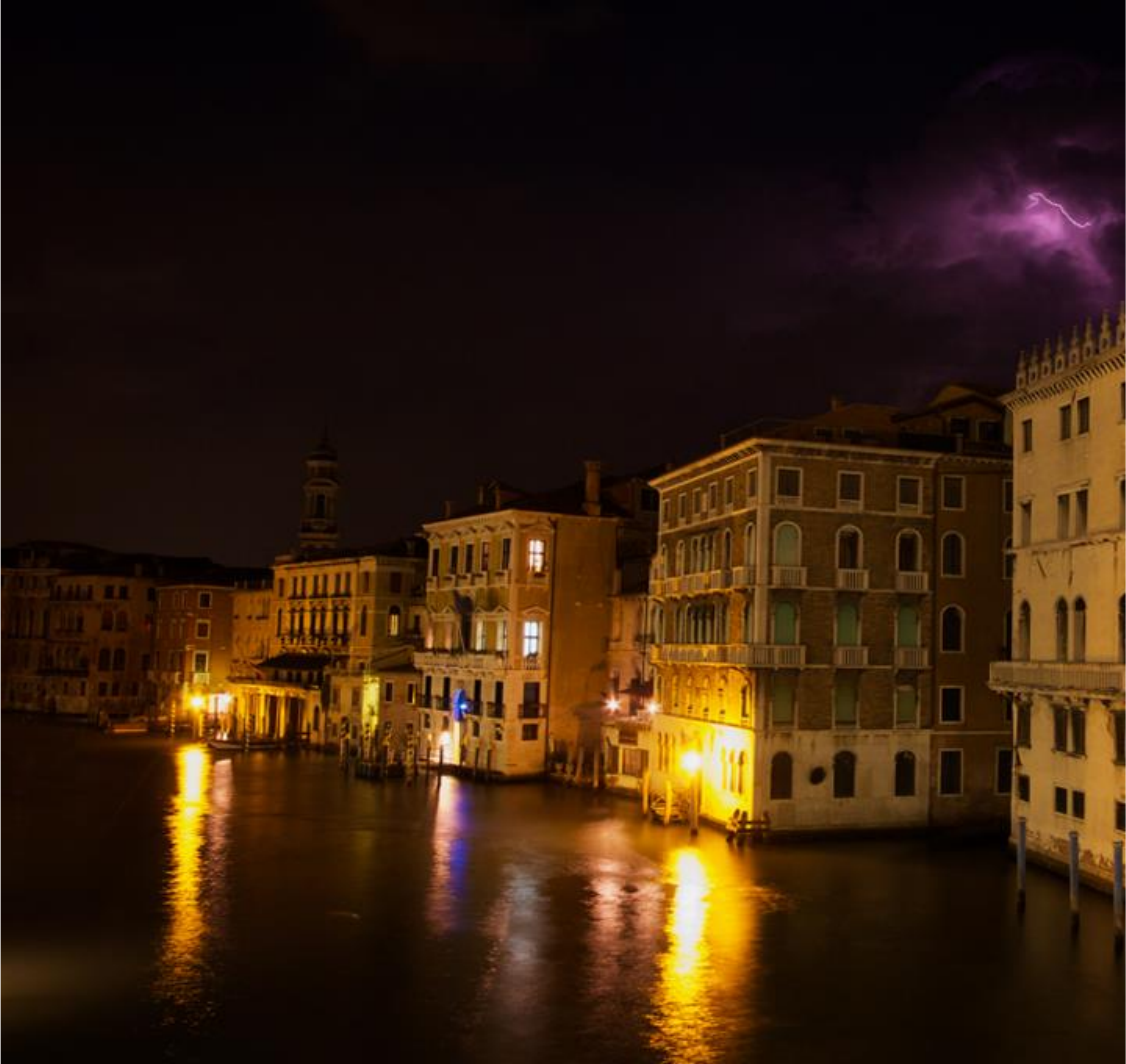} &
        \includegraphics[width=0.32\linewidth]{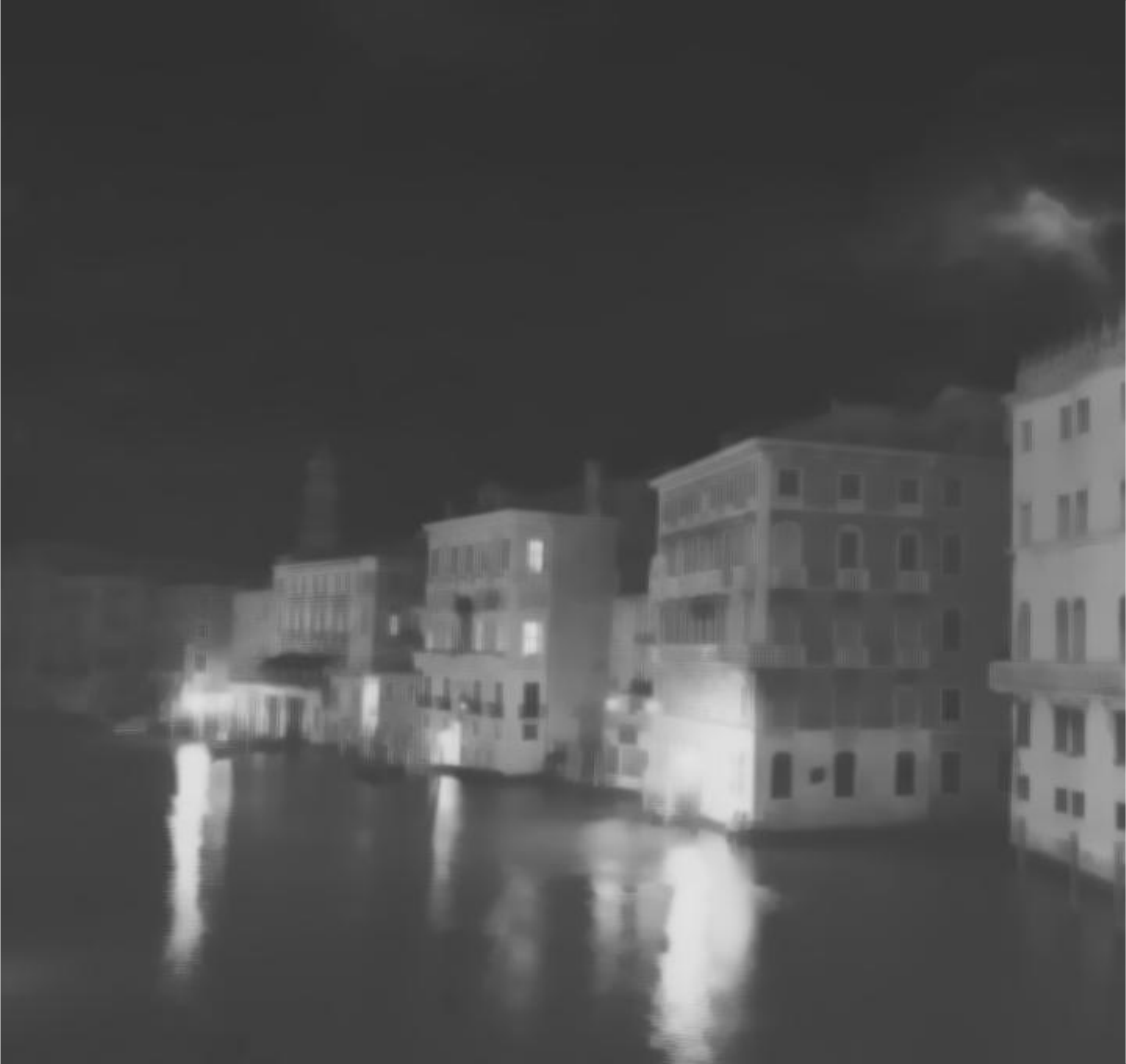} &
        \includegraphics[width=0.32\linewidth]{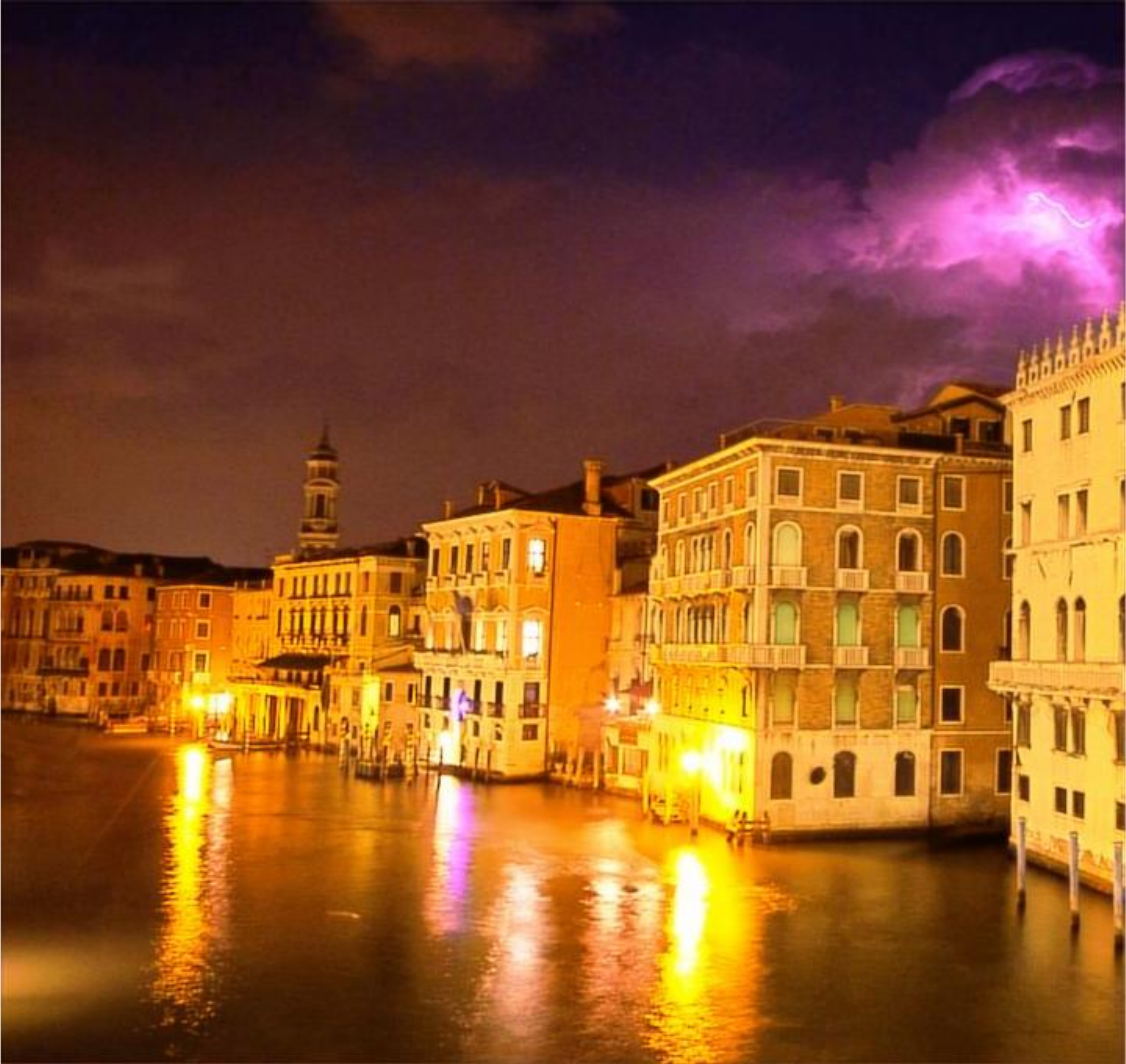} \\  
          (a) & (b) & (c)  \\
         \includegraphics[width=0.32\linewidth]{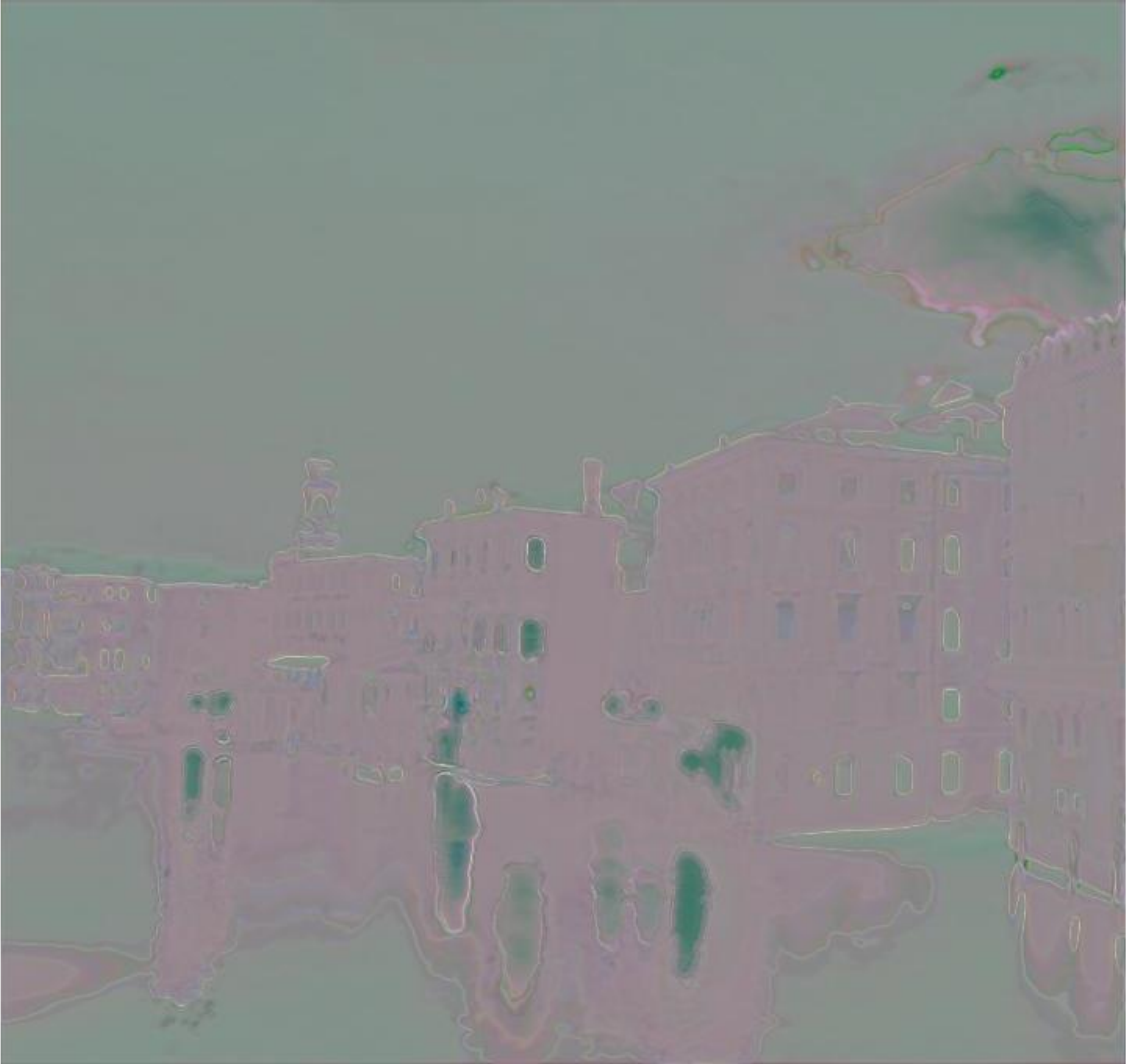}&
         \includegraphics[width=0.32\linewidth]{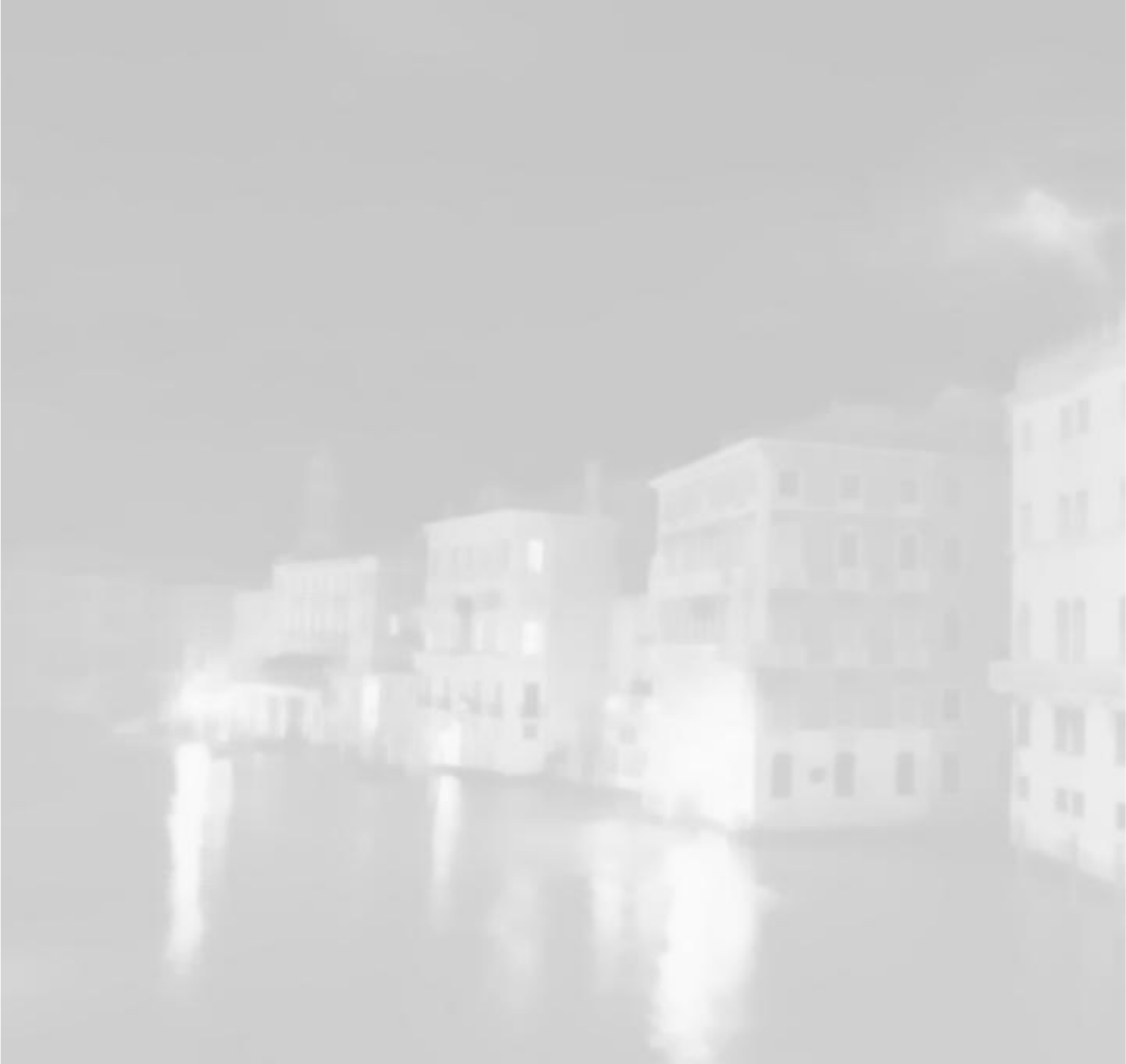}&
         \includegraphics[width=0.32\linewidth]{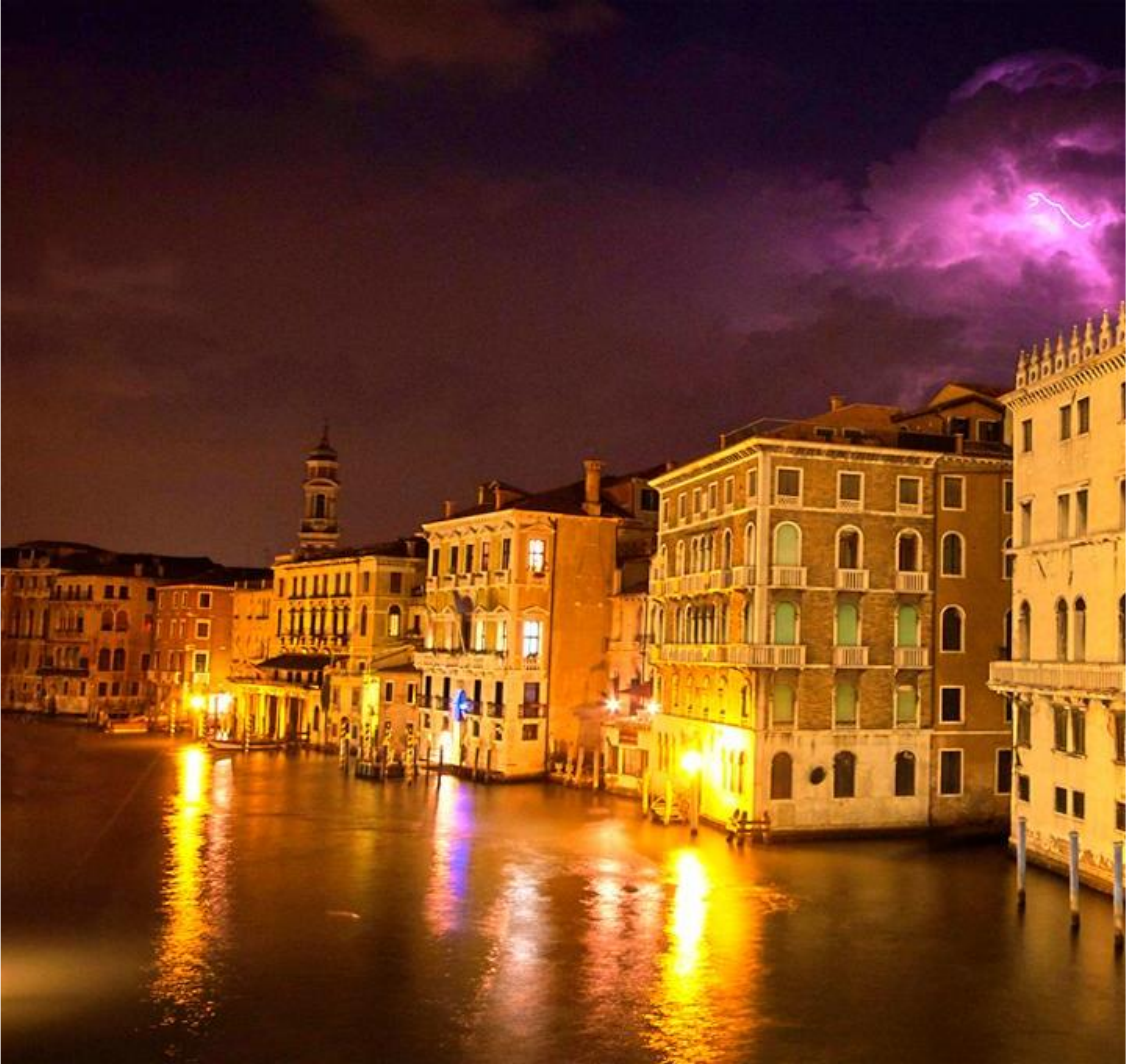} \\
         (d)& (e) & (f) \\            
    \end{tabular}
    \caption{The results of Retinex decomposition and denoise of ZERRINNet, (a) is the Original low-light image, (b), (c), (d), (e) and (f) represent its illumination, reflectance, noise-map, adjust-illumination and enhanced resul. Please zoom in to see the details. More results can be found in \href{https://github.com/liwenchao0615/ZERRINNet}{project website}}
    \label{figure:algorithms}
    \vspace{-3mm}
\end{figure}

In order to get better enhancement effect, we analyze the whole enhancement process, we get the reflection component, light component and noise component by decoupling, as shown in Fig.\ref{figure:algorithms}, ideal light maps are segmented and smooth and contain only the structural information of the image, the reflection component contains the most original color information of the building, and the first row of Fig. performs Retinex decomposition very well. There is noise in the low-light image, which is hidden in the shadow details, as shown in Fig.~\textcolor{red}{\ref{figure:algorithms}(d)}. Here contains the noise information of the low-light image, and more thorough denoising can be realized by simple decoupling.
%Furthermore, we explored the effect of the number of iterations on the enhancement results, firstly, we visualized the iteration process, as shown in Fig~\ref{fig:iter_result}. Through the experiments, we found that after the number of iterations reached 100, as shown in Fig~\ref{fig:iter_result2}. The evaluation indexes of the images (NIQE~\cite{1} and NIQMC~\cite{2}) did not change much, fluctuating above and below a certain fixed value, which provided a basis for the selection of the subsequent parameters.
%\begin{figure*}[!t]
 % \centering
 % \includegraphics[width=\textwidth]{images4/iter_result0.pdf}
 % \caption{Visualization of our enhancement method at different number of iterations, please zoom %in to see the details.}
 % \label{fig:iter_result}
%\end{figure*}

%\begin{figure}[!t]
%\centering
%\includegraphics[width=\linewidth]{images4/iter_rresult1g.pdf}
%\caption{ The relationship between the number of iterations and the evaluation metrics of the image, where$\uparrow$ for a high score value is the best, while$\downarrow$ representing a low score value is the best.}
%\label{fig:iter_result2}
%\end{figure}

\subsection{Experimental Details}
Our experiments were carried out on a single NVIDIA GeForce GTX 3080TI GPU using Pytorch.
For the parameter settings, the $\mathrm{\delta}$, $\mathrm{\gamma}$,  $\mathrm{\lambda}_i$, $\mathrm{\lambda}_k$, $\mathrm{\lambda}_n$ and the maximum iterations $\mathrm{M}$ are set as 0.1, 0.4, 2, 2, 6000 and 1000, iterative process using the ADAM optimizer with the parameters are $\beta_1 = 0.9$, $\beta_2 = 0.999$, and $\mathrm{l}_r$ = 0.001.

Furthermore, in order to evaluate the performance of the proposed method, we compare the proposed method with some of the best performing and typical low-light image enhancement methods, these methods include Retinex-Net\cite{retinex-net} , ExCNet\cite{ExCNet} , MBLLEN\cite{MBLLEN}, kinD++\cite{kind++} , Zero-DCE\cite{guo2020zero} , Zero-DCE++\cite{Zero-DCE++}, EnlightenGAN\cite{enlightengan}, RUAS\cite{RUAS}, SGZ\cite{SGZ}, SCI\cite{sci}, LLFlow\cite{llFlow}, URetinex-Net\cite{uretinex}. We refer to the official source code given for these methods and keep the same experimental environment with default parameters and model weights on a public low-light image image dataset.

\textit{1) Experimental Datasets:}
 In order to study the performance of the algorithms, we tested the method proposed in this paper and the method compared with it on nine low-light datasets, which contain seven public low-light datasets and two self-collected datasets of real photographed low-light images and datasets of low-light images synthesised by artificial synthesis. These two datasets are self-made real dataset and synthetic LOL dataset, which are mainly used to verify the generalization performance of the method. Which are briefly described below.

These seven public datasets are classified as DICM \cite{6467022}, ExDark \cite{loh2019getting}, Fusion\cite{fu2016fusion}, LIME\cite{7782813} , MEF\cite{7120119}, NPE\cite{6512558}, VV. 
 DICM \cite{6467022}contains 69 low-light images, the scenes include night, evening twilight and backlight, in this experiment, we chose the previous 44 images for the experiment. ExDark\cite{loh2019getting} consists of 14763 low-light images of ten different types, the scenes include low light, environment, object, single, weak, strong, screen, window, shadow and Twilight, divided into 12 object categories, and in this experiment we randomly selected 36 low-light images from these 12 categories. 
 Fusion \cite{fu2016fusion} Contains 16 low-light images of varying sizes.
 VV consists of 24 low-light images of individual lives, the LIME dataset \cite{7782813} provides 10 low-light images. The Fusion dataset collects 18 low-light image datasets, and the NPE dataset \cite{6512558} contains 84 images with cloudy daytime, daybreak, nightfall and nighttime scenes, 8 low-light images were selected for our experiments, and the MEF dataset \cite{7120119} contains 17 images with scenes including nighttime and twilight.
 
\textit{2) Performance Criteria:} In order to evaluate the performance of the proposed method in low light image enhancement, we used three unsupervised low light image quality assessment metrics, namely Natural image quality evaluator (NIQE) \cite{1}, no reference image quality metric for contrast distortion (NIQMC)\cite{2} and the colorfulness-based patch-based contrast quality index (CPCQI) \cite{gu2016no}. NIQE \cite{1} is an assessment of the overall naturalness of the enhanced image, with lower values reflecting a higher overall naturalness of the image. CPCQI \cite{gu2016no}is an assessment of the enhancement effect between the input and the enhanced output in terms of average intensity, signal strength, and signal structure components, and measures the contrast quality of the image, with larger values conveying higher contrast. NIQMC\cite{2} is the value of contrast distortion, with larger values reflecting an image with higher high quality.

\IEEEpubidadjcol
\section{EXPERIMENTAL EVALUATION}
In this section, the performance of the proposed enhancement method is first evaluated from the subjective and objective viewpoints of the image. Immediately after that, we test the generalization performance of our method, both in real low-light datasets and in advanced visual tasks, and finally, we perform ablation experiments to further explore the role of each part of our method.
\subsection{Algorithm Performance Analysis}
In this subsection, in comparison with the current state-of-the-art methods, we will demonstrate the effectiveness of our method in terms of subjective and objective evaluations.

\begin{figure*}[!tbp]
    \centering
    \small
    \tabcolsep=0.05cm
    \begin{tabular}{c c c c c c c}
        \includegraphics[width=0.142\linewidth]{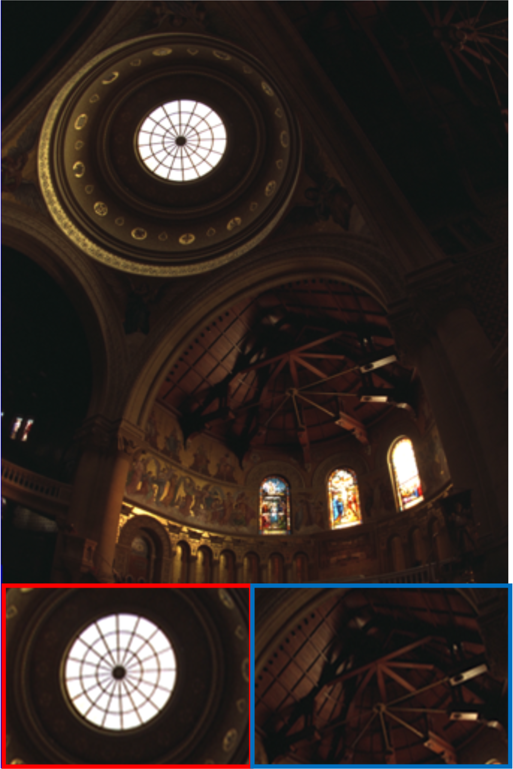} &
        \includegraphics[width=0.142\linewidth]{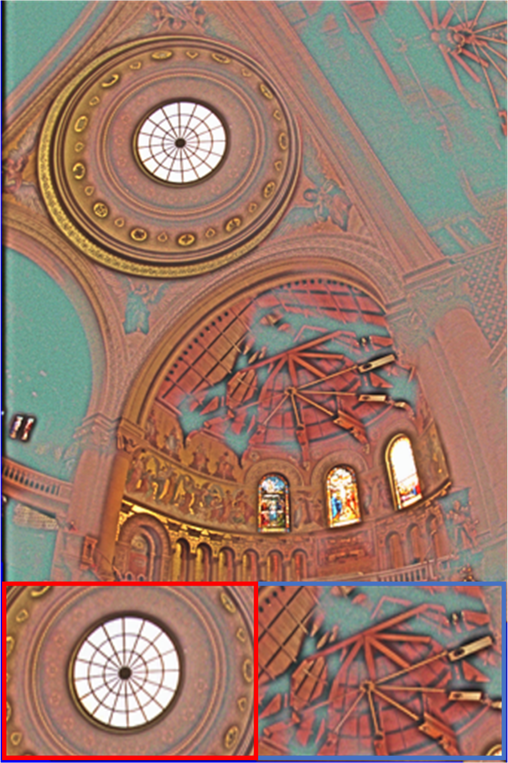} &
        \includegraphics[width=0.142\linewidth]{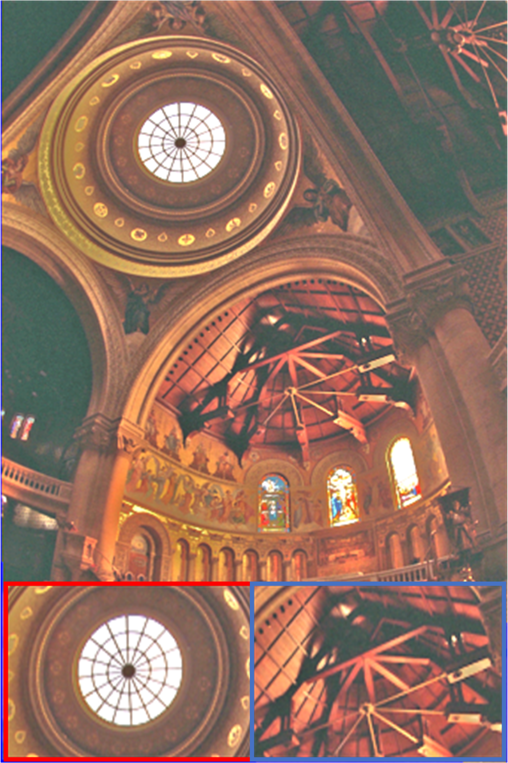}&
         \includegraphics[width=0.142\linewidth]{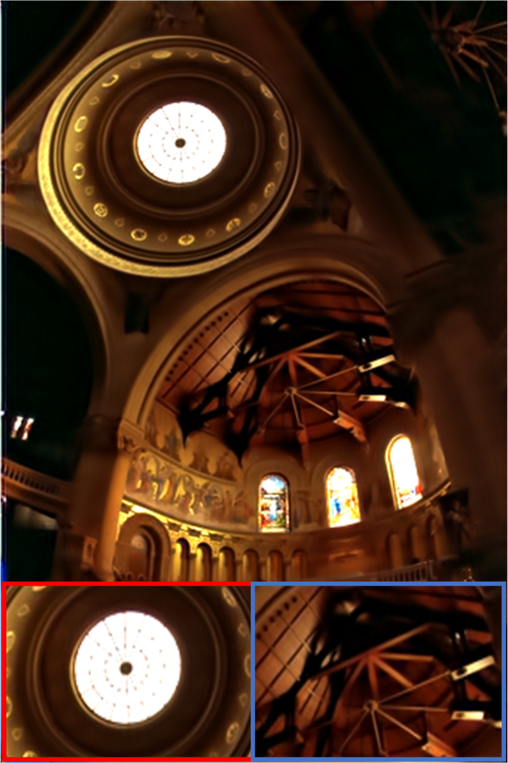}&
         \includegraphics[width=0.142\linewidth]{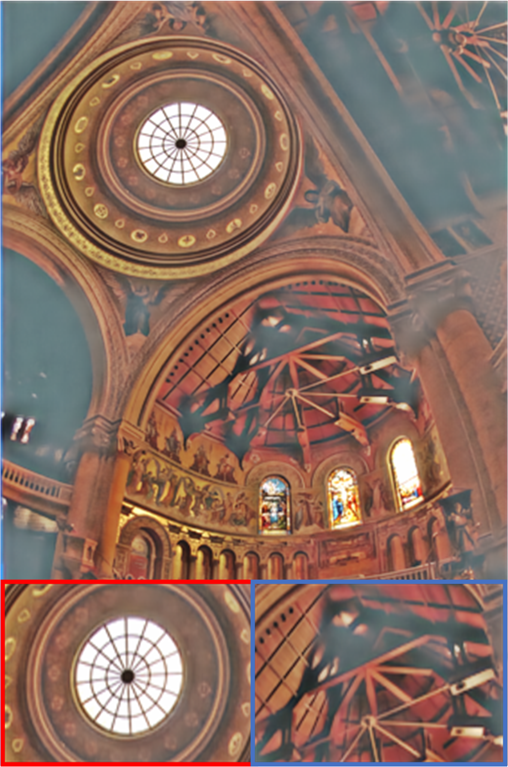}&
         \includegraphics[width=0.142\linewidth]{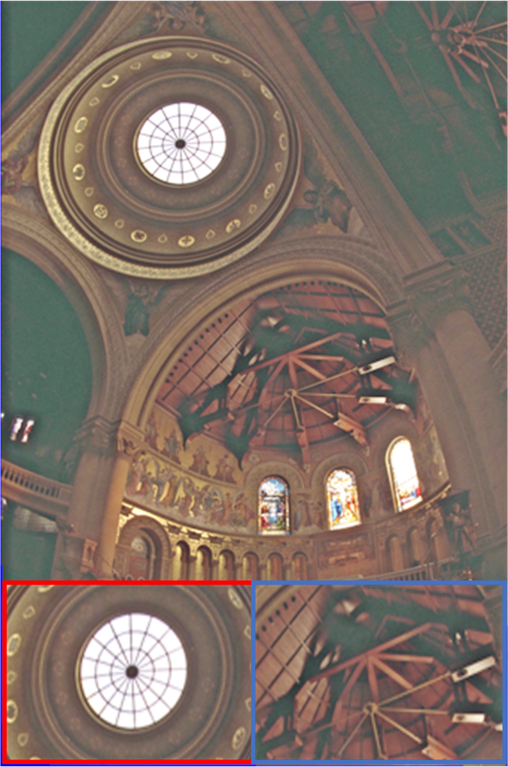}&
         \includegraphics[width=0.142\linewidth]{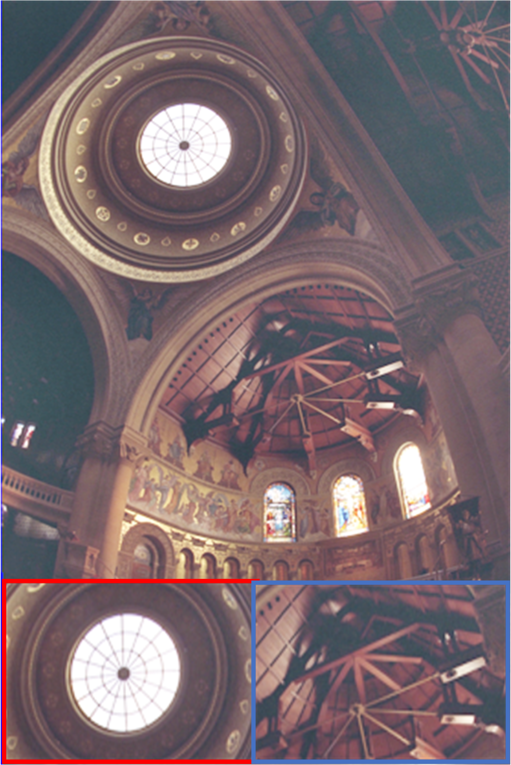} \\  
          (a)input & (b)Retinex-Net & (c)ExCNet & 
          (d)MBLLEN & (e)kinD++ & (f)Zero-DCE & (g)Zero-DCE++  \\
         \includegraphics[width=0.142\linewidth]{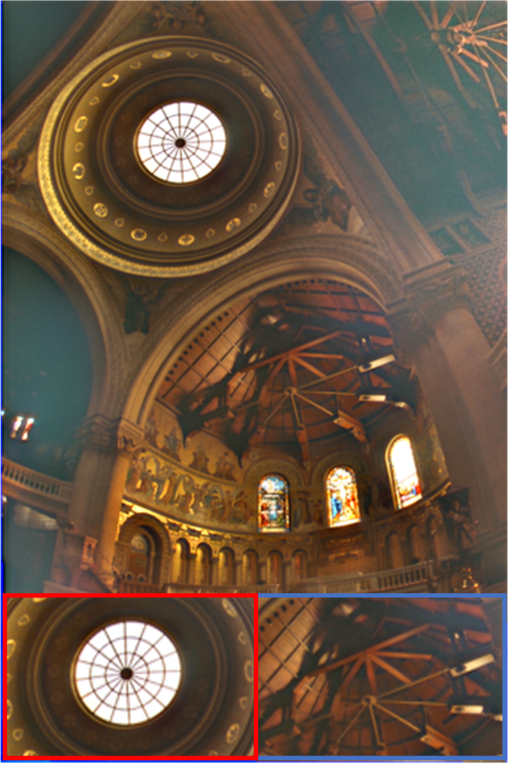}&
         \includegraphics[width=0.142\linewidth]{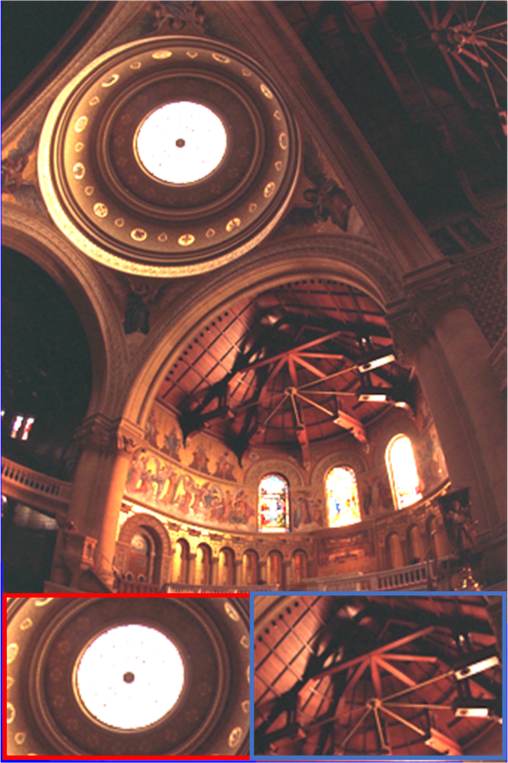}&
         \includegraphics[width=0.142\linewidth]{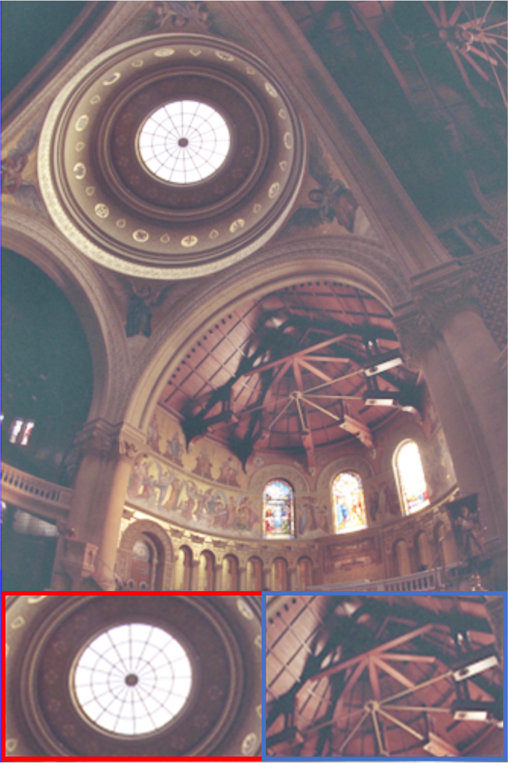}&
         \includegraphics[width=0.142\linewidth]{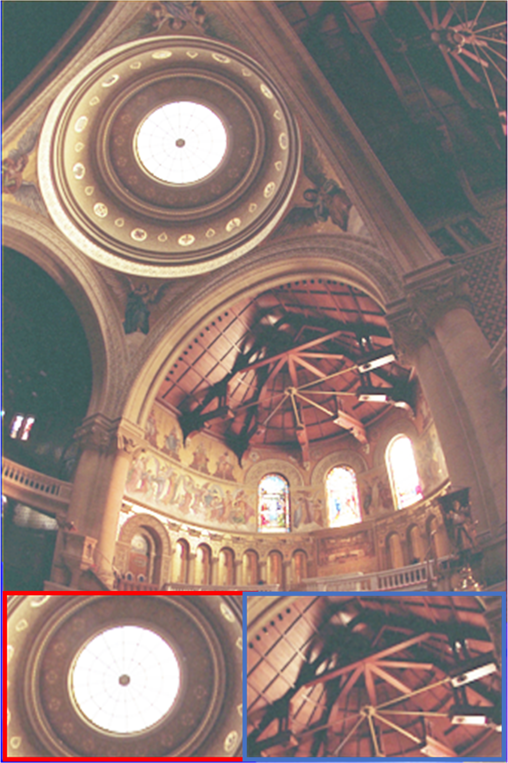}&
         \includegraphics[width=0.142\linewidth]{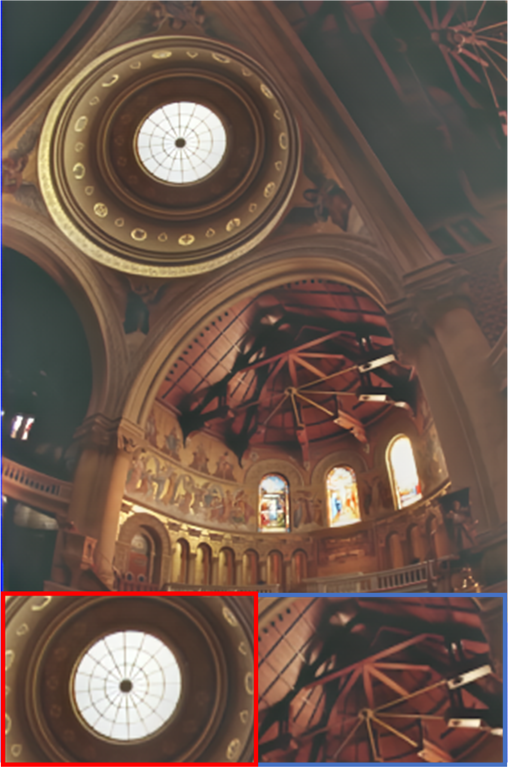}&
         \includegraphics[width=0.142\linewidth]{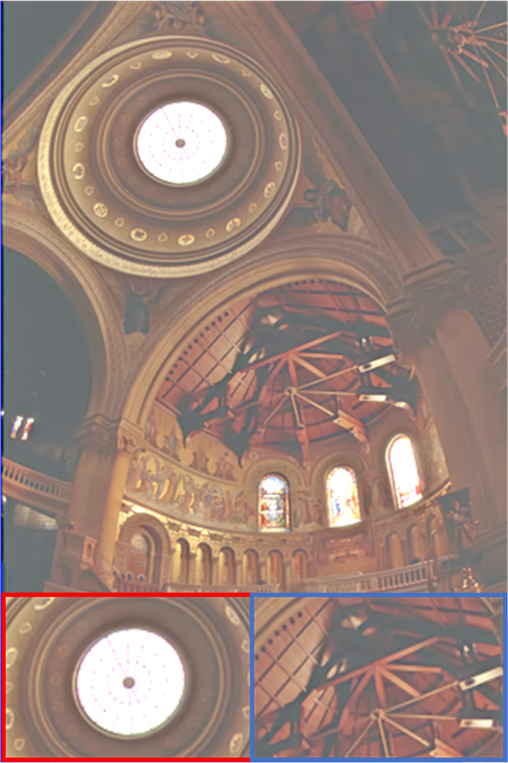}&
         \includegraphics[width=0.142\linewidth]{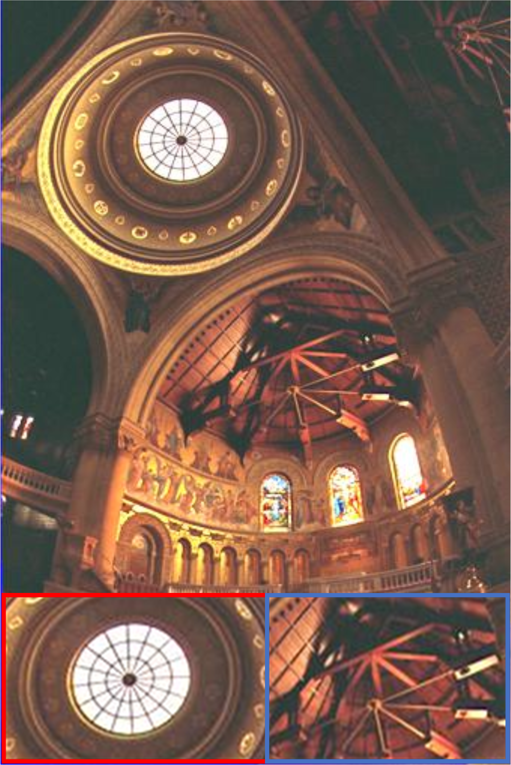} \\
         (h)EnlightenGAN & (i)RUAS & (j)SGZ &  (k)SCI &
         (l)LLFlow & (m)URetinex-Net & (n)ZERRINNet\\   
         
    \end{tabular}
    \caption{ Compared to other methods, our results perfectly recover image details and have better contrast. You can zoom in to see the details, and more results for all the datasets tested by these methods can be found in \href{https://github.com/liwenchao0615/ZERRINNet}{project website}. These methods include Retinex-Net\cite{retinex-net} , ExCNet\cite{ExCNet} ,MBLLEN\cite{MBLLEN}, kinD++\cite{kind++} , Zero-DCE\cite{Zero-DCE} , Zero-DCE++\cite{Zero-DCE++}, EnlightenGAN\cite{enlightengan}, RUAS\cite{RUAS}, SGZ\cite{SGZ}, SCI\cite{sci}, LLFlow\cite{llFlow}, URetinex\cite{uretinex} and ZERRINNet.}
    \label{fig: 4}
    \vspace{-3mm}
\end{figure*}

\begin{figure*}[!tbp]
    \centering
    \tabcolsep=0.05cm
    \begin{tabular}{c c c c c c c}
        \includegraphics[width=0.14\linewidth]{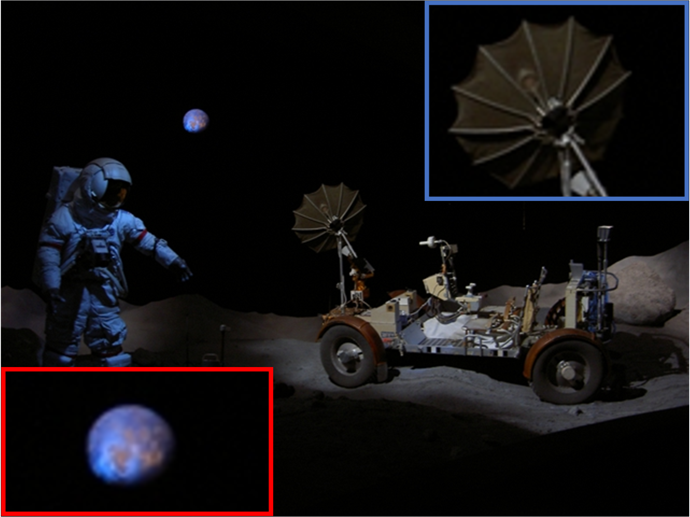} &
        \includegraphics[width=0.14\linewidth]{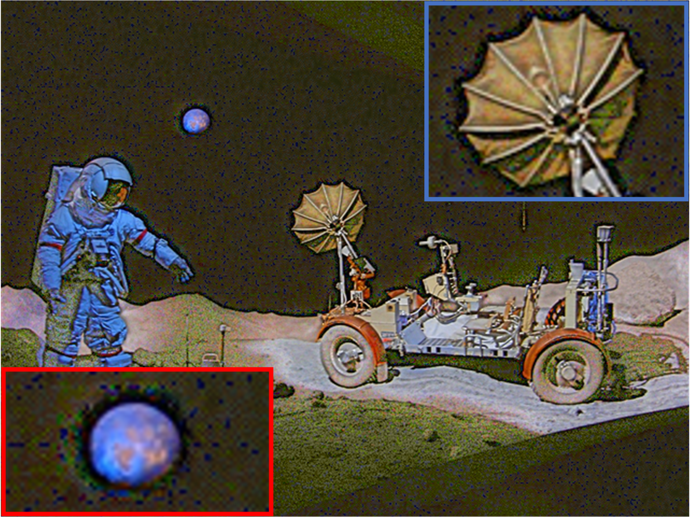} &
        \includegraphics[width=0.14\linewidth]{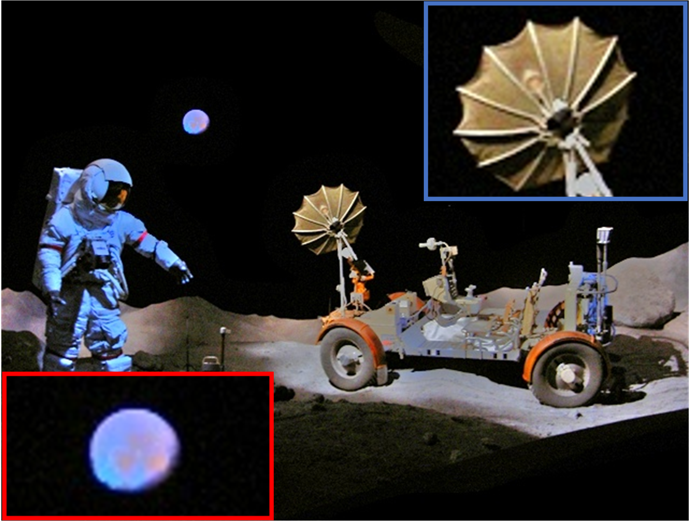}&
         \includegraphics[width=0.14\linewidth]{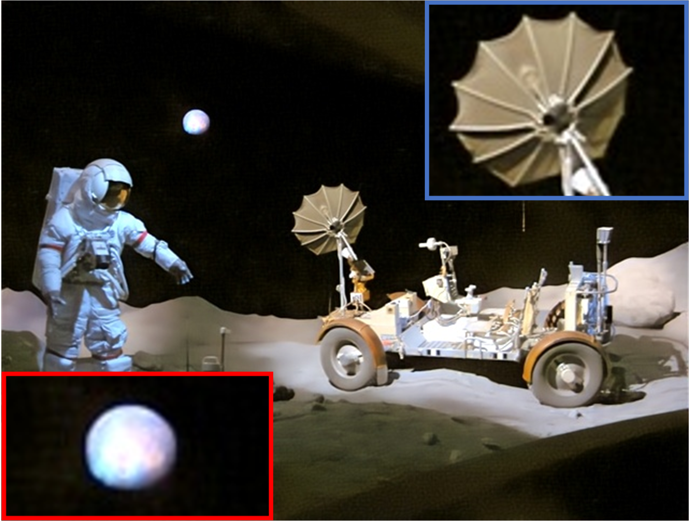}&
         \includegraphics[width=0.14\linewidth]{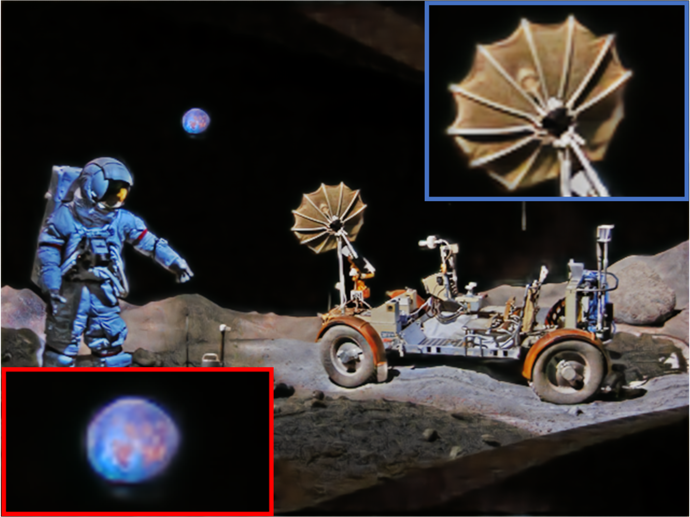}&
         \includegraphics[width=0.14\linewidth]{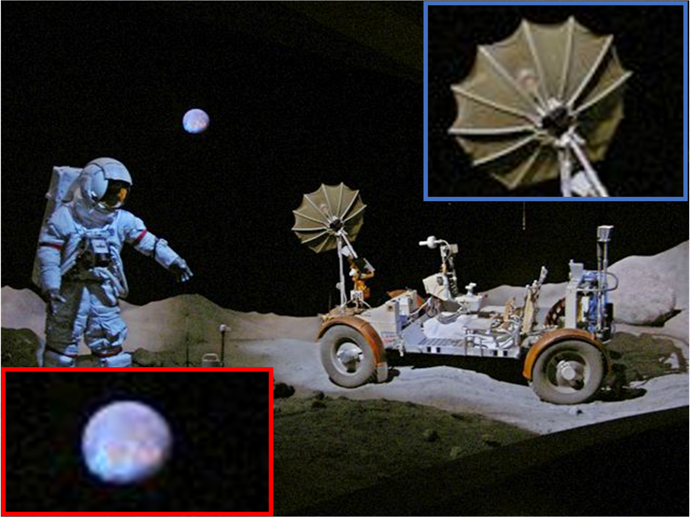}&
         \includegraphics[width=0.14\linewidth]{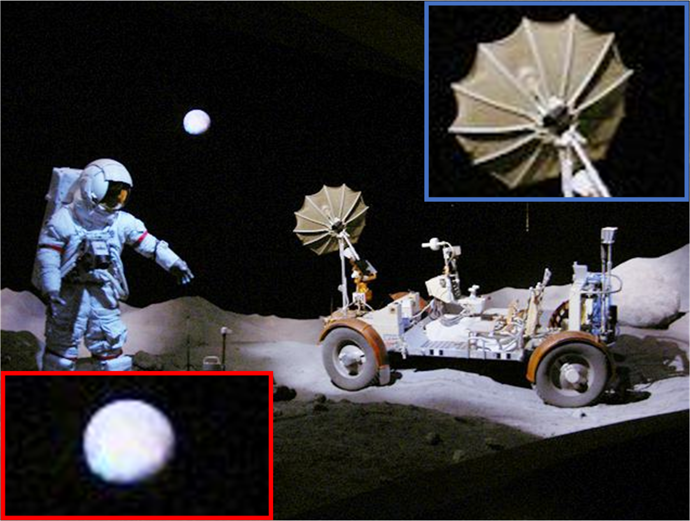} \\  
          (a)input & (b)Retinex-Net & (c)ExCNet & 
          (d)MBLLEN & (e)kinD++ & (f)Zero-DCE & (g)Zero-DCE++   \\
         \includegraphics[width=0.14\linewidth]{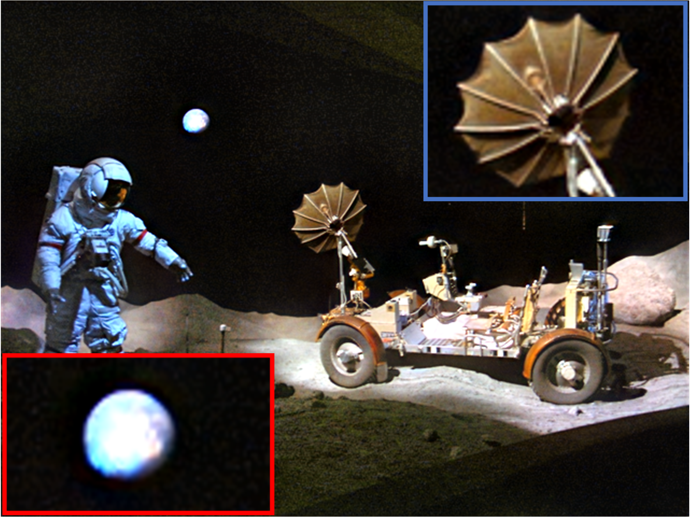}&
         \includegraphics[width=0.14\linewidth]{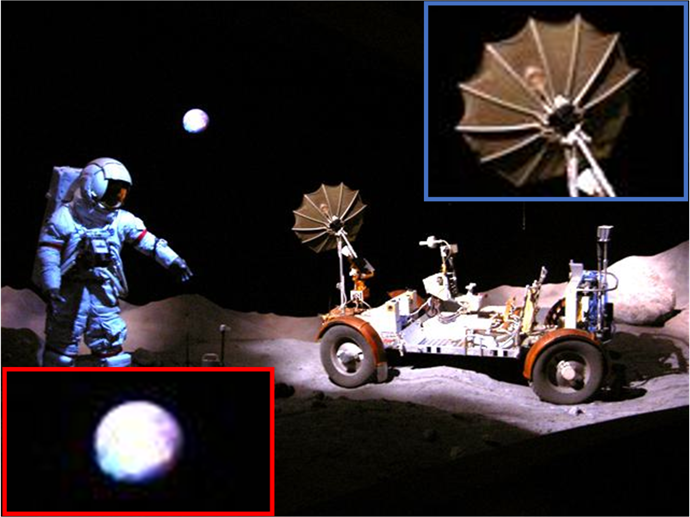}&
         \includegraphics[width=0.14\linewidth]{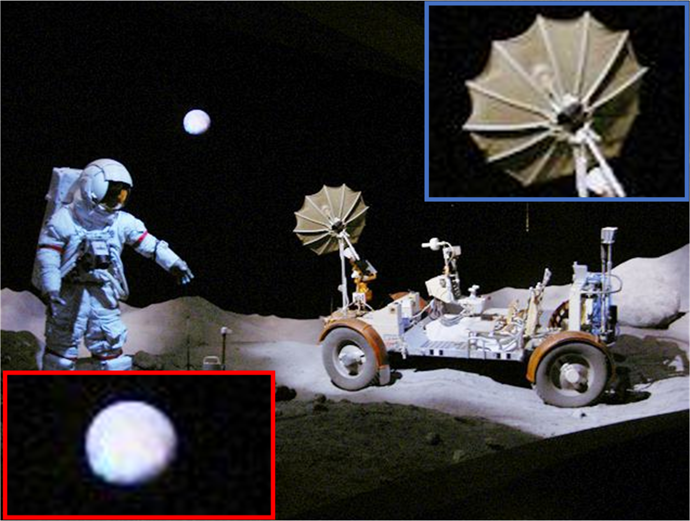}&
         \includegraphics[width=0.14\linewidth]{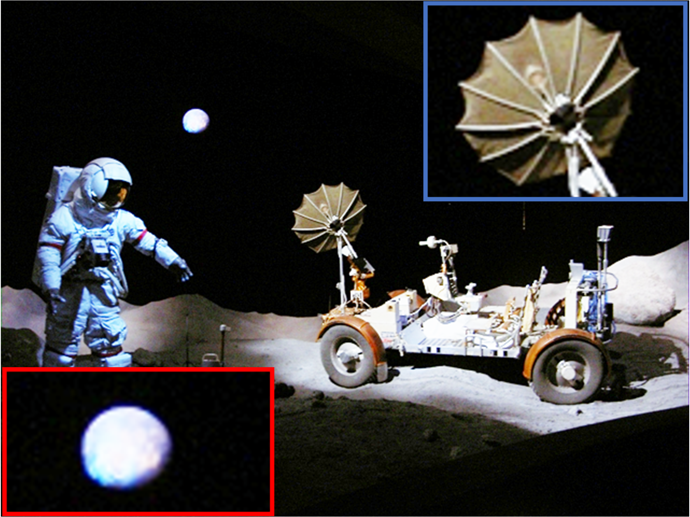}&
         \includegraphics[width=0.14\linewidth]{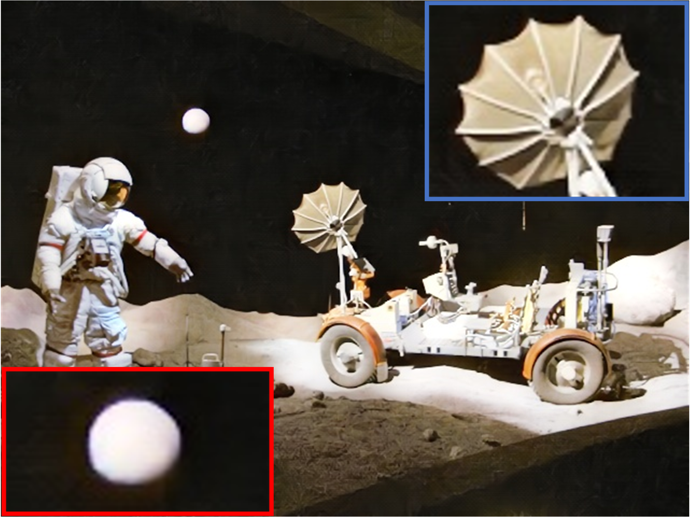}&
         \includegraphics[width=0.14\linewidth]{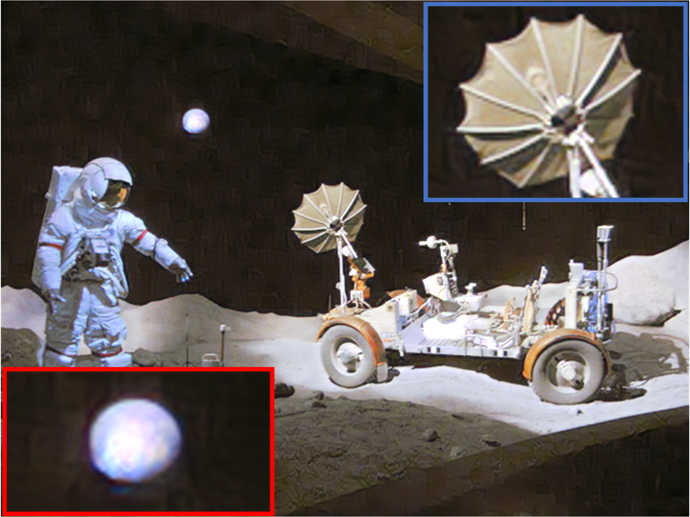}&
         
         \includegraphics[width=0.14\linewidth]{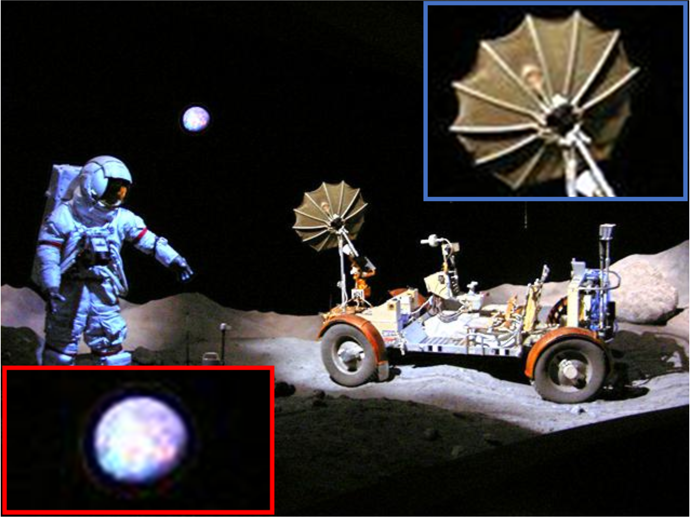} \\
         (h)EnlightenGAN & (i)RUAS & (j)SGZ &  (k)SCI &
         (l)LLFlow & (m)URetinex-Net & (n)ZERRINNet   \\
         
    \end{tabular}
    \caption{ Compared to other methods, it can be noticed that our results have competitive contrast and more natural images. You can zoom in to see the details.}
    \label{fig: 5}
    \vspace{-3mm}
\end{figure*}

\begin{figure*}[!tbp]
    \centering
    \tabcolsep=0.05cm
    \begin{tabular}{c c c c c c c}
        \includegraphics[width=0.14\linewidth]{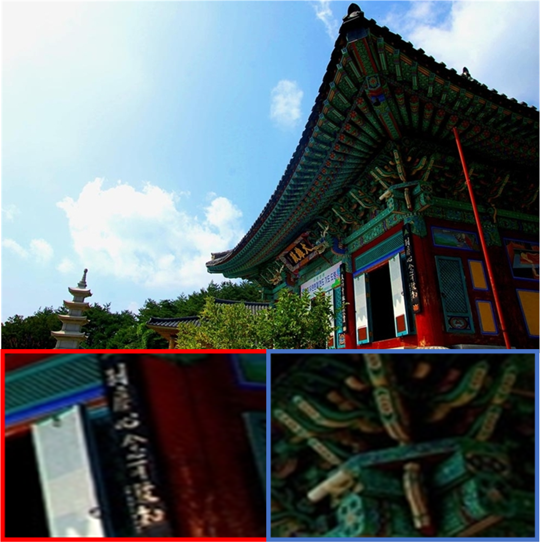} &
        \includegraphics[width=0.14\linewidth]{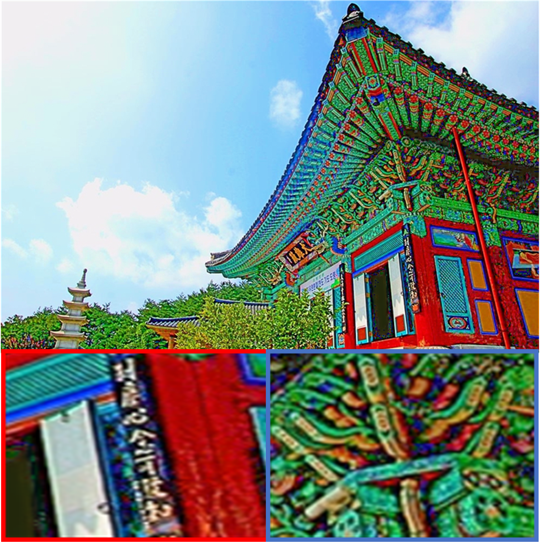} &
        \includegraphics[width=0.14\linewidth]{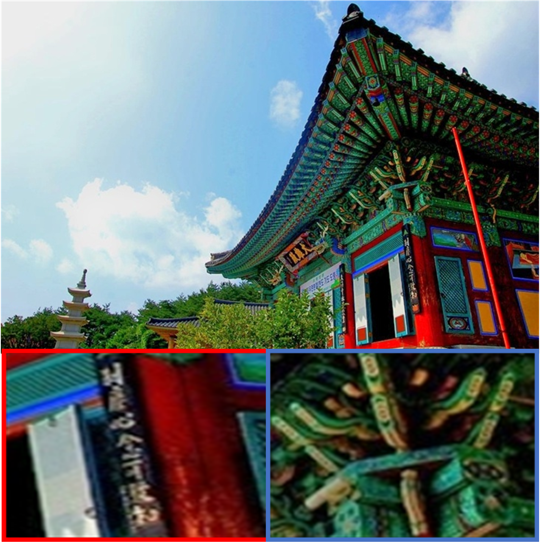}&
         \includegraphics[width=0.14\linewidth]{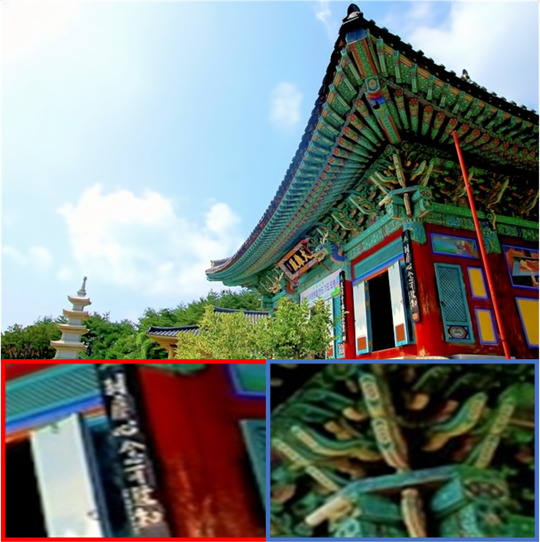}&
         \includegraphics[width=0.14\linewidth]{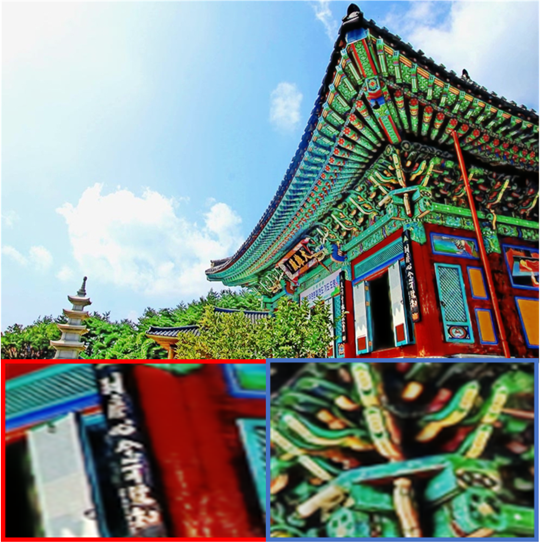}&
         \includegraphics[width=0.14\linewidth]{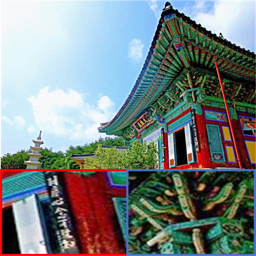}&
         \includegraphics[width=0.14\linewidth]{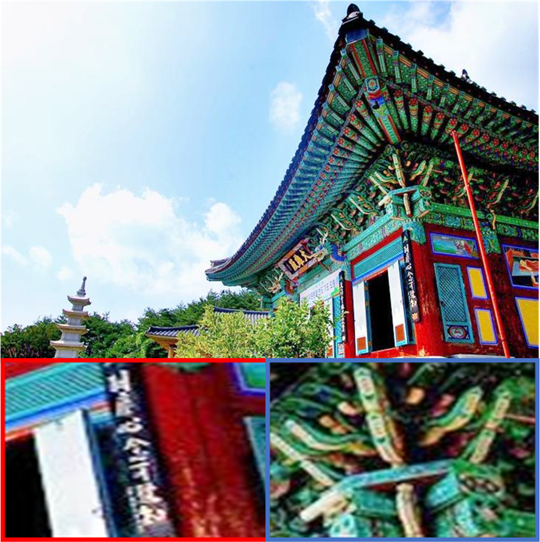} \\  
          (a)input & (b)Retinex-Net & (c)ExCNet & 
          (d)MBLLEN & (e)kinD++ & (f)Zero-DCE & (g)Zero-DCE++   \\
         \includegraphics[width=0.14\linewidth]{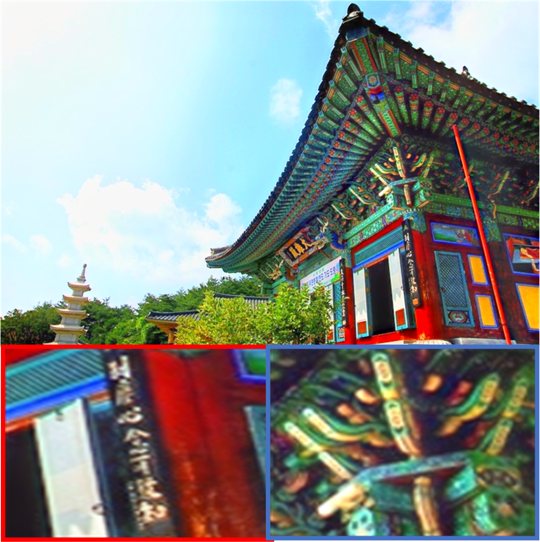}&
         \includegraphics[width=0.14\linewidth]{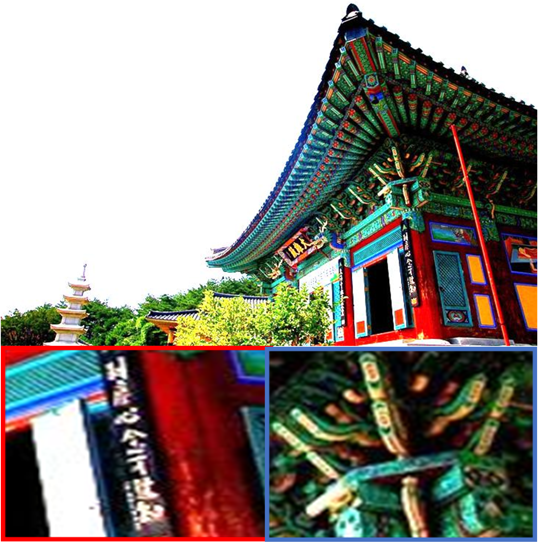}&
         \includegraphics[width=0.14\linewidth]{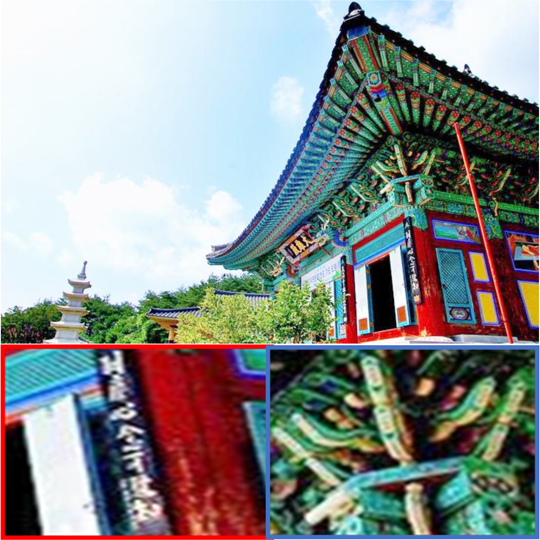}&
         \includegraphics[width=0.14\linewidth]{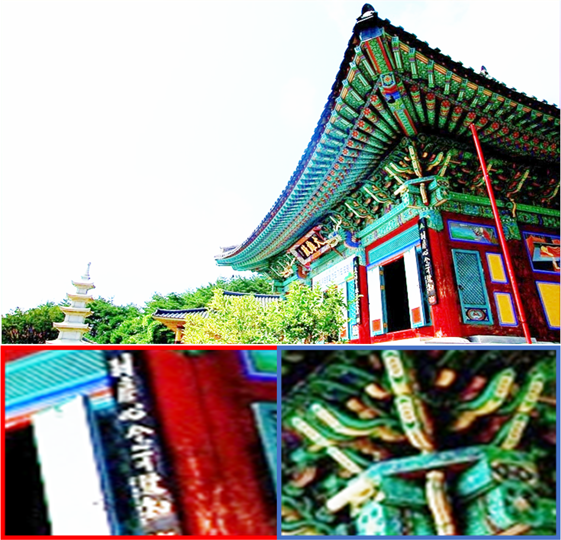}&
         \includegraphics[width=0.14\linewidth]{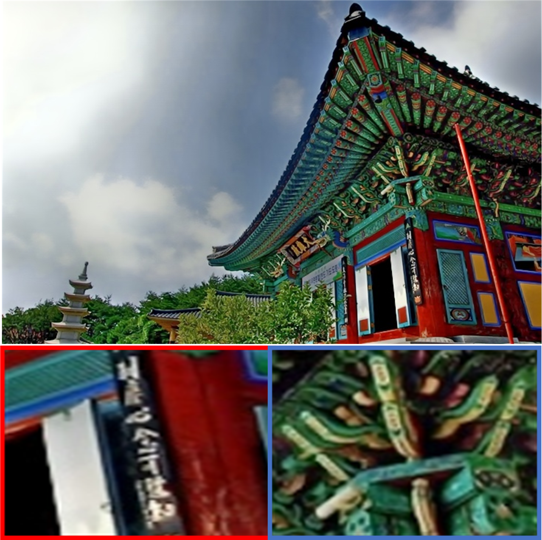}&
         \includegraphics[width=0.14\linewidth]{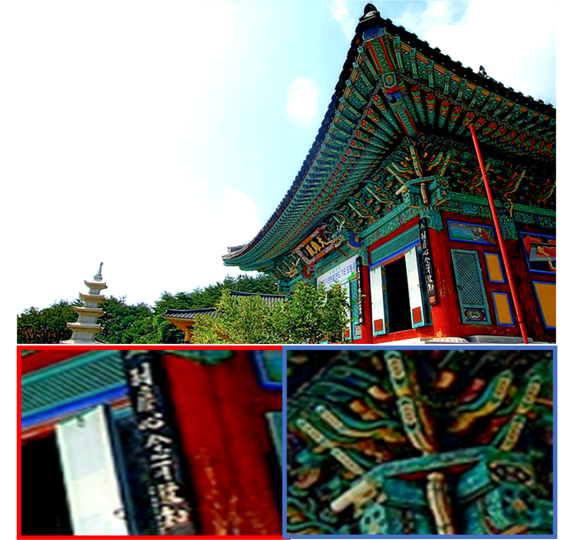}&
         \includegraphics[width=0.14\linewidth]{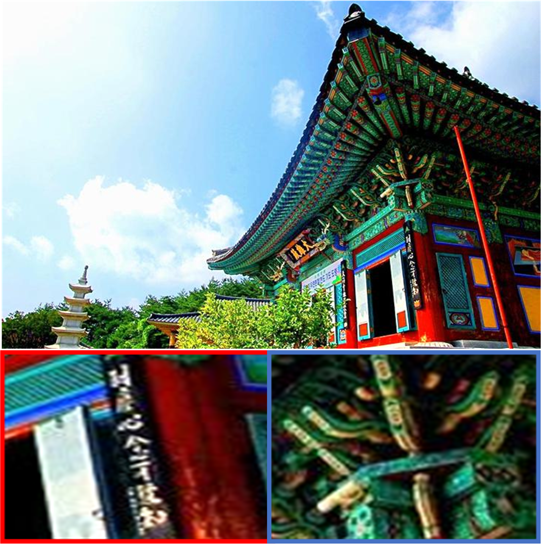} \\
         (h)EnlightenGAN & (i)RUAS & (j)SGZ &  (k)SCI &
         (l)LLFlow & (m)URetinex-Net & (n)ZERRINNet   \\
         
    \end{tabular}
    \caption{ Compared to other methods, it can be noticed that no color distortion and better visual quality of images occur in our results. You can zoom in to see the details.}
    \label{fig: 6}
    \vspace{-3mm}
\end{figure*}

\textit{1) Quantitative assessments:}
As stated above, we have compared our proposed method with the state-of-the-art methods at this stage on these publicly available datasets and to ensure the fairness of our experiments, we have used the official code without any processing. For the visual comparison, we have chosen the typical enhancement images of several different methods, as shown in Figs \ref{fig: 4}, \ref{fig: 5}, and \ref{fig: 6}. First of all, we carefully observed the enhancement results corresponding to each enhancement method in Fig. \ref{fig: 4}, and we found that the overall visual perception of our method is the best, first of all, in terms of the noise in the dark region of the image, except for RUAS, MBLLEN, and our method, all other methods have varying degrees of noise in the dark region, with Retinex-Net being the most fatal and ExCNet Relatively less symptomatic, these noises mask the detailed information in the dark region of the image. In addition, RUAS, MBLLEN and our method have better contrast of the image, unfortunately, there is overexposure in the bright region of the image, which makes the details of the bright region lost, in order to facilitate the observation, we specially zoom in this part of the labeling to show that, while our method in the bright region of the information is full and clear. In the end, our method is more popular in terms of overall visual perception.

In addition, in Fig. \ref{fig: 5}, we examined the contrast and naturalness of the image, and our proposed method ZERRINNet not only enhances the contrast of the image, but also perfectly maintains the naturalness of the image, and LLFlow and URetinex-Net sacrifice some details despite improving the overall brightness of the image, and it can also be seen from the figure that these two methods and the previously mentioned MBLLEN all show different degrees of color distortion. Finally, as shown in Fig. \ref{fig: 6}, we also compare our method with other methods in images underexposed during the day, and the results show that RUAS, SCI, and URetinex-Net show serious information loss, and the original blue-sky background is directly turned into a white background, while the LLFlow method shows color distortion, and compared to other methods, our method still maintains a strong competitive edge compared to other methods. The above only lists the three scenarios of outdoor, night and daytime, in the experiments, we have done a lot of comparative experiments, including all the images of the above-mentioned datasets, in addition, we have also added other methods, and these enhancement results will be revealed in the open source code repository, which will be convenient for the research of the scholars in the future.

\textit{2) Qualitative assessments:}
In this section, we present a quantitative comparison of our method with the best performing method at this stage. As mentioned above, we evaluate our enhancement results using the values of three unsupervised evaluation metrics, NIQE, NIQMC and CPCQI, which reflect the overall performance of image enhancement. The results of the experiments are shown in Tables \ref{tab:niqe}, \ref{tab:CPCQI} and \ref{tab:NIQMC}, for visualization, we especially use red, green, and blue fonts to represent the top three.
First, as shown in Table \ref{tab:niqe}, our ZERRINNet has the lowest NIQE average, which means that our method can effectively maintain the naturalness of the image and give a better visual impression. Zero-DCE++ and SGZ are in the second and third place, respectively, which we illustrate in terms of principle.
\begin{table*}[!t]
\small
\centering
\caption{
COMPARISON OF AVERAGE \textbf{QINE}{$\downarrow$} ON SEX DATASETS}
\label{tab:niqe}
\begin{tabular}{@{}lcccccccc@{}}
\toprule
\textbf{Method} & \textbf{DICM}                 & \textbf{ExDark}               & \textbf{Fusion}               & \textbf{LIME} & \textbf{MEF}                  & \textbf{VV} & \textbf{NPE} & \multicolumn{1}{l}{\textbf{Average}} \\ \midrule
RetinexNet   & 4.5001                        & 5.0959                        & 4.1652                        & 4.5906                        & 4.3949                        & {\color[HTML]{FE0000} 2.6933} & 4.5648                        & 4.2864                        \\
ExCNet       & 3.6243                        & 4.1277                        & 3.6012                        & {\color[HTML]{34FF34} 3.7745} & {\color[HTML]{34FF34} 3.3595} & 2.7976                        & 4.0955                        & 3.6258                        \\
kinD++       & 3.1165                        & {\color[HTML]{34FF34} 3.9783} & 3.5994                        & 4.7216                        & 3.7365                        & {\color[HTML]{34FF34} 2.7016} & 4.3796  & 3.7476                        \\
Zero-DCE++   & {\color[HTML]{34FF34} 2.8871} & 4.2328                        & 3.5168                        & 3.971                         & 3.3967                        & 3.0947                        & {\color[HTML]{FE0000} 4.0231} & {\color[HTML]{34FF34} 3.5789} \\
EnlightenGAN & {\color[HTML]{3531FF} 2.9427} & {\color[HTML]{3531FF} 4.0099} & {\color[HTML]{FE0000} 3.3821} & {\color[HTML]{FE0000} 3.6585} & {\color[HTML]{FE0000} 3.219}  & 4.1281                        & 4.1104                        & 3.6358                        \\
RUAS         & 4.0802                        & 4.4603                        & 4.6601                        & 4.2463                        & 3.8297                        & 4.6137                        & 5.5342                        & 4.4892                        \\
SGZ          & {\color[HTML]{FE0000} 2.874}  & 4.237                         & {\color[HTML]{3531FF} 3.4899} & 3.9556                        & {\color[HTML]{3531FF} 3.3857} & 3.1413                        & {\color[HTML]{34FF34} 4.0443} & {\color[HTML]{3531FF} 3.5896} \\
SCI          & 3.6418                        & 4.639                         & 3.9073                        & 4.1802                        & 3.6312                        & 2.9171                        & 4.4705                        & 3.9124                        \\
LLFlow       & 3.7952                        & {\color[HTML]{FE0000} 3.7215} & 3.6648                        & 4.1507                        & 3.7318                        & {\color[HTML]{3531FF} 2.7471} & {\color[HTML]{000000} 4.399}  & 3.6351                        \\
URetinex-Net & 3.4587                        & 4.1289                        & 4.0097                        & 4.3395                        & 3.7891                        & 3.0189                        & 4.6928                        & 3.9197                        \\ \midrule
ZERRINNet     & 3.417                         & 4.0199                        & {\color[HTML]{34FF34} 3.4465} & {\color[HTML]{3531FF} 3.8987} & 3.4528                        & 2.7879                        & {\color[HTML]{3531FF} 4.0907} & {\color[HTML]{FE0000} 3.5876} \\ \bottomrule
\end{tabular}
\end{table*}
Zero-DCE++ is to fit the luminance mapping curve by neural network and use this curve to brighten the image, and at the same time reduce the computational overhead by channel-by-channel divisible convolution, however, as to the authors, the idea only focuses on the luminance of the image, image blurring and details, and does not solve it very well.The basic idea of SGZ is to augment the low-light image by semantic guidance. The local details of the image can be effectively recovered, for which it is reasonable to understand that these two methods have good performance. Compared with these methods, our ZERRINNet can effectively address the challenges faced by low-light images. In addition, the CPCOI value is a comprehensive reflection of the quality of an image, including contrast, sharpness, brightness, color and naturalness, and in Table \ref{tab:CPCQI}, we have equally achieved the best score in the CPCOI average. Meanwhile, our method is the highest on almost all datasets, except (VV and LIME datasets), ExCNet and SCI are ranked second and third, while the SGZ score is the lowest. Finally, as can be seen in Table \ref{tab:NIQMC}, our method is still highly competitive, although second in the ranking and LLFow first in the ranking, in LLFlow first it is a supervised low-light image enhancement method, which requires pairs of datasets to be trained, and in addition the method does not take into account color distortions, as can be seen from the enhancement results of the qualitative analysis in Fig \ref{fig: 6}.

% \usepackage{booktabs}
\begin{table*}[!t]
\small
\centering
\caption{
COMPARISON OF AVERAGE \textbf{CPCQI}{$\uparrow$} ON SEVEN DATASETS}
\label{tab:CPCQI}
\begin{tabular}{@{}lclclclcl@{}}
\toprule
\textbf{Method} & \textbf{DICM}                 & \textbf{ExDark}               & \textbf{Fusion}               & \textbf{LIME} & \textbf{MEF}                  & \textbf{VV} & \textbf{NPE} & \multicolumn{1}{l}{\textbf{Average}} \\ \midrule
RetinexNet      & 0.9232                        & 0.9532                              & 0.5513                        & 0.8007                            & 0.8863                        & {\color[HTML]{3531FF} 0.916}    & {\color[HTML]{3531FF} 0.9187} & 0.8499                               \\
ExCNet          & {\color[HTML]{34FF34} 1.0037} & 0.8404                              & {\color[HTML]{FE0000} 0.9957} & {\color[HTML]{FE0000} 1.0817}     & {\color[HTML]{34FF34} 1.0555} & {\color[HTML]{FE0000} 0.9906}   & {\color[HTML]{34FF34} 0.9987} & {\color[HTML]{34FF34} 0.9951}        \\
KinD++          & 0.9144                        & 0.7119                              & 0.4314                        & 0.6053                            & 0.8603                        & {\color[HTML]{34FF34} 0.9585}   & 0.6414                        & 0.7319                               \\
Zero-DCE++      & 0.5657                        & 0.3488                              & 0.4951                        & 0.3799                            & 0.3817                        & 0.8824                          & 0.4033                        & 0.4938                               \\
EnlightenGAN    & {\color[HTML]{3531FF} 0.9656} & {\color[HTML]{34FF34} 0.9753}       & 0.5528                        & 0.8071                            & 0.9751                        & 0.7669                          & 0.8887                        & 0.8473                               \\
RUAS            & 0.7956                        & 0.9419                              & 0.7929                        & {\color[HTML]{34FF34} 1.0791}     & {\color[HTML]{3531FF} 0.9792} & 0.7975                          & 0.7951                        & 0.8830                               \\
SGZ             & 0.5393                        & 0.3355                              & 0.4686                        & 0.3689                            & 0.3654                        & 0.8322                          & 0.3731                        & 0.4690                               \\
SCI             & 0.9007                        & {\color[HTML]{3531FF} 0.9569}       & 0.8105                        & {\color[HTML]{3531FF} 1.0293}     & 0.949                         & 0.8088                          & 0.7616                        & {\color[HTML]{3531FF} 0.8881}        \\
LLFlow          & 0.8395                        & 0.8922                              & {\color[HTML]{3531FF} 0.8893} & 0.9143                            & 0.8949                        & 0.8451                          & 0.8671                        & 0.8774                               \\
URetinex-Net    & 0.8369                        & 0.8793                              & 0.5163                        & 0.7493                            & 0.8175                        & 0.7869                          & 0.862                         & 0.7783                               \\
kinD            & 0.8627                        & 0.8642                              & 0.566                         & 0.7995                            & 0.9096                        & 0.8954                          & 0.9485                        & 0.8351                               \\ \midrule
ZERRINNet       & {\color[HTML]{FE0000} 1.1033} & {\color[HTML]{FE0000} 1.2348}       & {\color[HTML]{34FF34} 0.9581} & 0.9714                            & {\color[HTML]{FE0000} 1.1617} & 0.5145                          & {\color[HTML]{FE0000} 1.0948} & {\color[HTML]{FE0000} 1.0055}        \\ \bottomrule
\end{tabular}
\end{table*}
% \usepackage{booktabs}
\begin{table*}[!t]
\small
\centering
\caption{
COMPARISON OF AVERAGE \textbf{NIQMC}{$\uparrow$} ON SEVEN DATASETS}
\label{tab:NIQMC}
\begin{tabular}{@{}lclclclcl@{}}
\toprule
\textbf{Method} & \textbf{DICM}                 & \textbf{ExDark}               & \textbf{Fusion}               & \textbf{LIME} & \textbf{MEF}                  & \textbf{VV} & \textbf{NPE} & \multicolumn{1}{l}{\textbf{Average}} \\ \midrule
RetinexNet      & 4.9073                        & 4.6591                              & 5.0026                        & 4.6771                            & 4.7473                        & 5.0779                          & 4.9669                        & 4.8626                               \\
ExCNet          & 5.1126                        & 4.9406                              & 5.0904                        & 5.1346                            & 4.9915                        & 5.3105                          & 5.0831                        & 5.0947                               \\
KinD++          & 5.1576                        & 4.8522                              & 5.2373                        & 4.882                             & 5.0592                        & 5.394                           & 5.1868                        & 5.1098                               \\
Zero-DCE++      & {\color[HTML]{34FF34} 5.3028} & 5.1206                              & 5.2461                        & 5.1277                            & 5.1719                        & {\color[HTML]{3531FF} 5.4488}   & {\color[HTML]{3531FF} 5.2784} & 5.2423                               \\
EnlightenGAN    & 5.252                         & 5.0881                              & {\color[HTML]{3531FF} 5.2462} & 5.0868                            & 5.1153                        & 5.4449                          & 5.152                         & 5.1979                               \\
RUAS            & 4.7747                        & 5.039                               & 4.5659                        & 4.9653                            & 4.7345                        & {\color[HTML]{34FF34} 4.4842}   & 4.7234                        & 4.7552                               \\
SGZ             & {\color[HTML]{3531FF} 5.2803} & 5.1624                              & 5.1876                        & {\color[HTML]{3531FF} 5.2066}     & {\color[HTML]{3531FF} 5.2151} & 5.4437                          & 5.2632                        & {\color[HTML]{3531FF} 5.2512}        \\
SCI             & 5.066                         & {\color[HTML]{3531FF} 5.2277}       & 4.8362                        & {\color[HTML]{FE0000} 5.3627}     & 5.1749                        & 4.9637                          & 5.1418                        & 5.1104                               \\
LLFlow          & {\color[HTML]{FE0000} 5.3819} & {\color[HTML]{FE0000} 5.2383}       & {\color[HTML]{FE0000} 5.3618} & {\color[HTML]{34FF34} 5.2547}     & {\color[HTML]{FE0000} 5.3355} & 5.4214                          & {\color[HTML]{34FF34} 5.3399} & {\color[HTML]{FE0000} 5.3333}        \\
URetinex-Net    & 5.1653                        & 5.0687                              & 5.0246                        & 5.0152                            & 5.0853                        & 5.1434                          & 5.2609                        & 5.1090                               \\ \midrule
ZERRINNet       & 5.0788                        & {\color[HTML]{34FF34} 5.2324}       & {\color[HTML]{34FF34} 5.2849} & 5.1683                            & {\color[HTML]{34FF34} 5.3087} & {\color[HTML]{FE0000} 5.5502}   & {\color[HTML]{FE0000} 5.4806} & {\color[HTML]{34FF34} 5.3005}    \\ \bottomrule   
\end{tabular}
\end{table*}
\subsection{Study of Generalisation Ability}
In this section, we fully validate the generalization performance of the proposed method from the tasks of shimmer image enhancement, face detection, target detection and instance segmentation.

\begin{figure*}[!htbp]
\centering
\includegraphics[width=\textwidth]{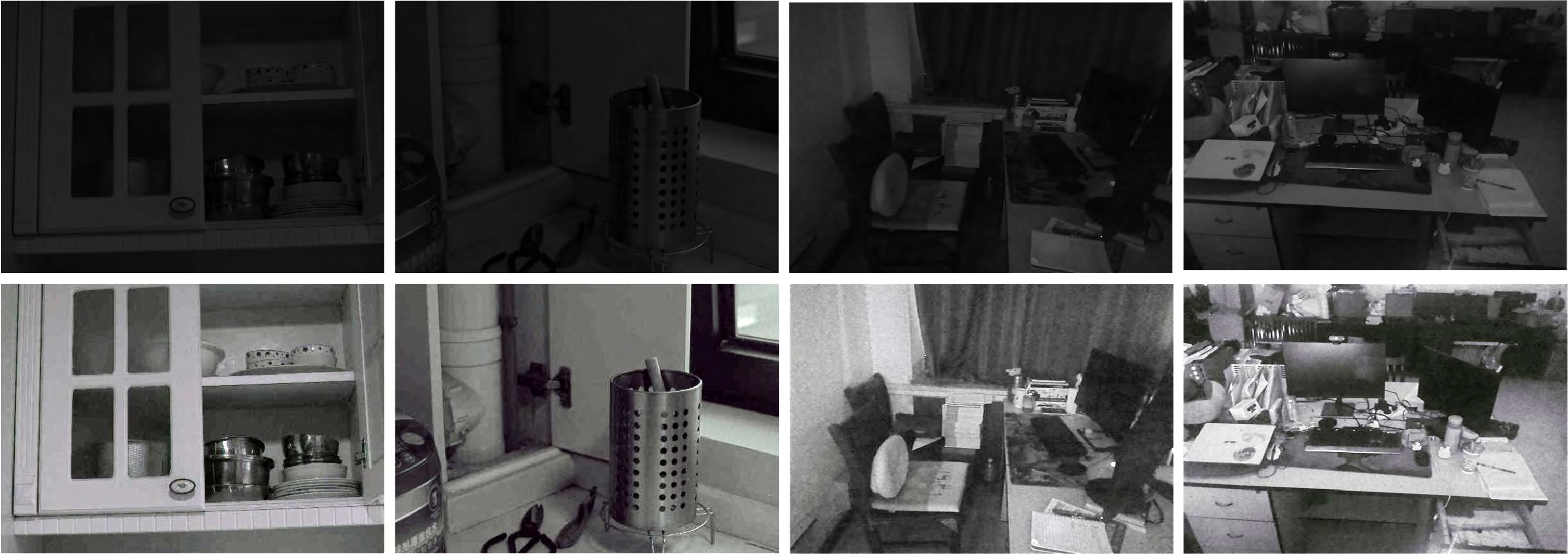}
\caption{Monochrome low-light image enhancement results, (two rows on the left) monochrome low-light photographic composite image, (two rows on the right) monochrome low-light photographic real shot low-light image.}
\label{greyscale:figure}
\end{figure*}

\begin{figure*}[htbp]
    \centering
    \includegraphics[width=\textwidth]{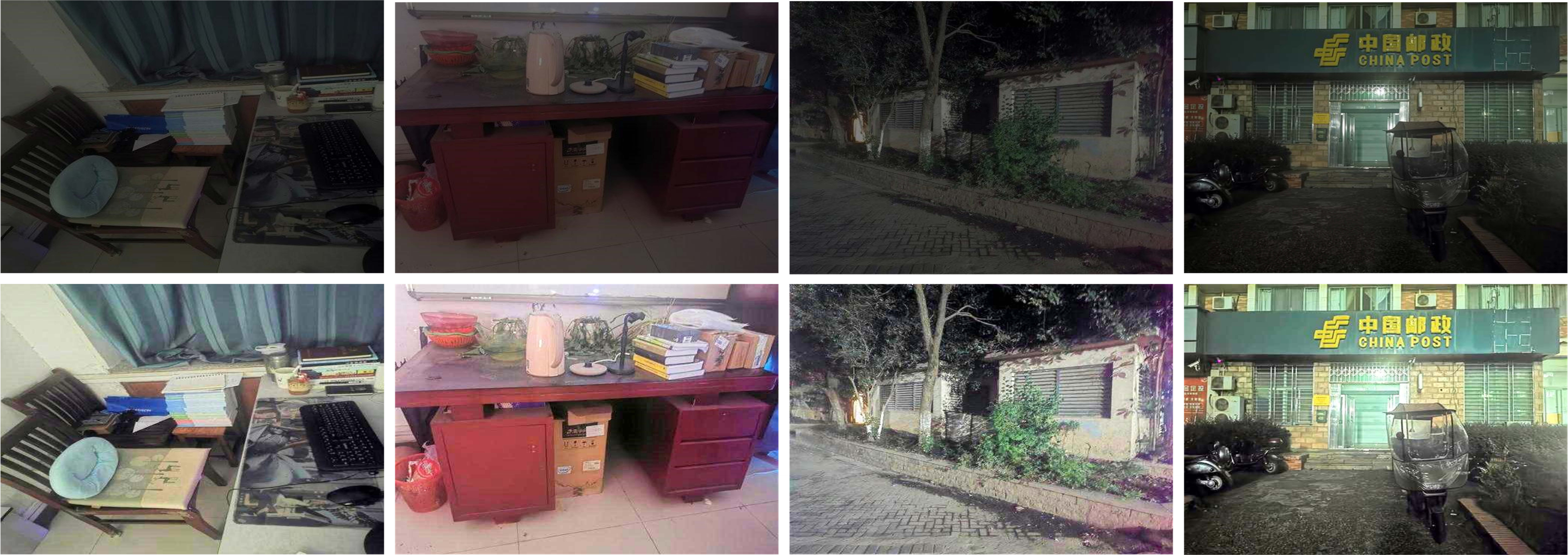}
    \caption{Color low-light image enhancement results, (two rows on the left) low-light photomontage images from the public dataset, and (two rows on the right) low-light images taken from a real night scene.}
    \label{colour:figure}
\end{figure*}

\begin{figure*}[!tbp]
    \centering
    \tabcolsep=0.05cm
    \begin{tabular}{c c }
        \includegraphics[width=0.5\linewidth]{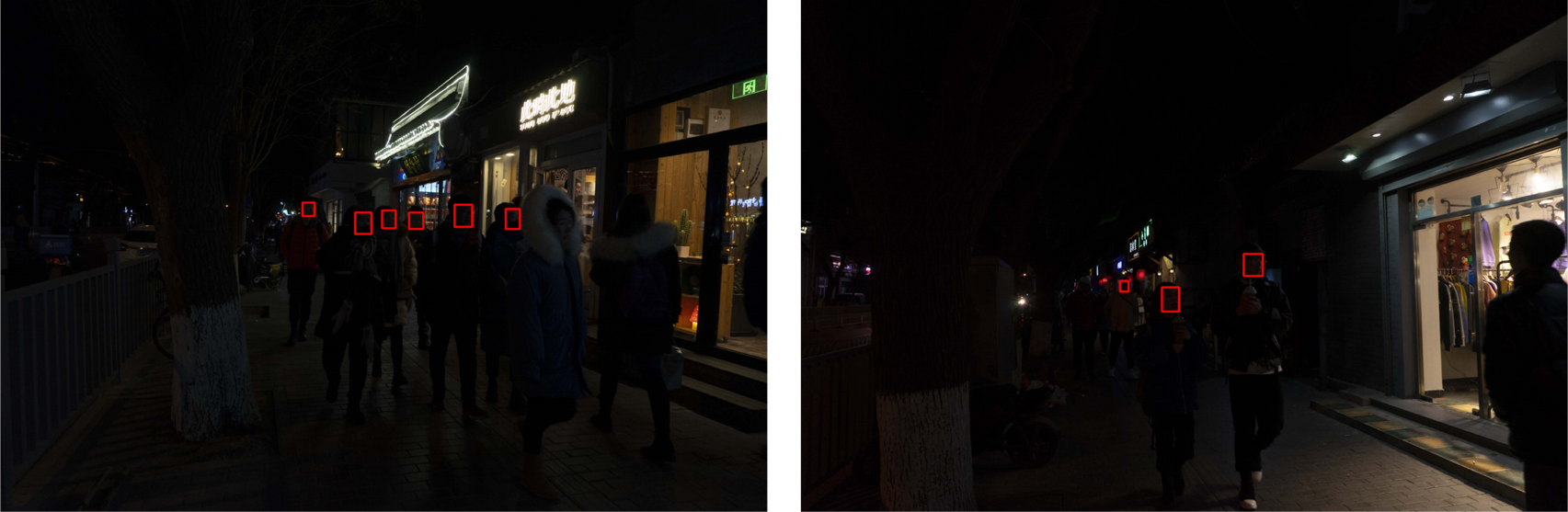}  &
        \includegraphics[width=0.5\linewidth]{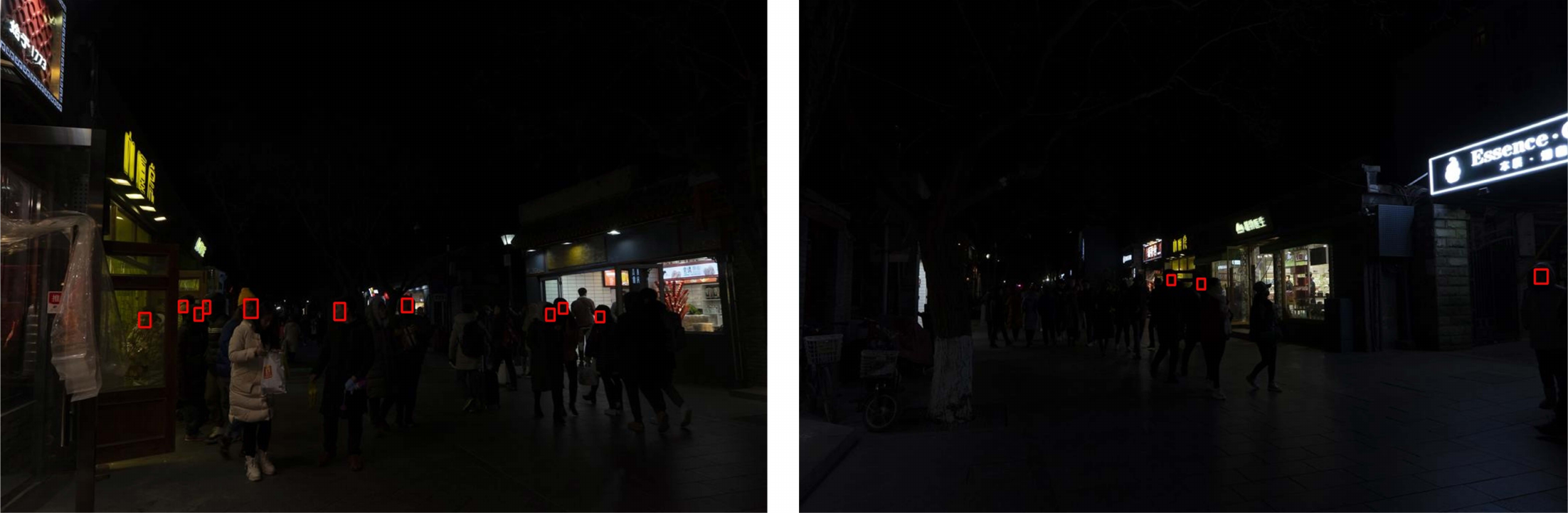} \\
         {\small (a) RetinaFace\cite{deng2020retinaface} + Low-light image}                          &
        {\small (b) DSFD \cite{li2019dsfd}+ Low-light image}                           \\
        \includegraphics[width=0.5\linewidth]{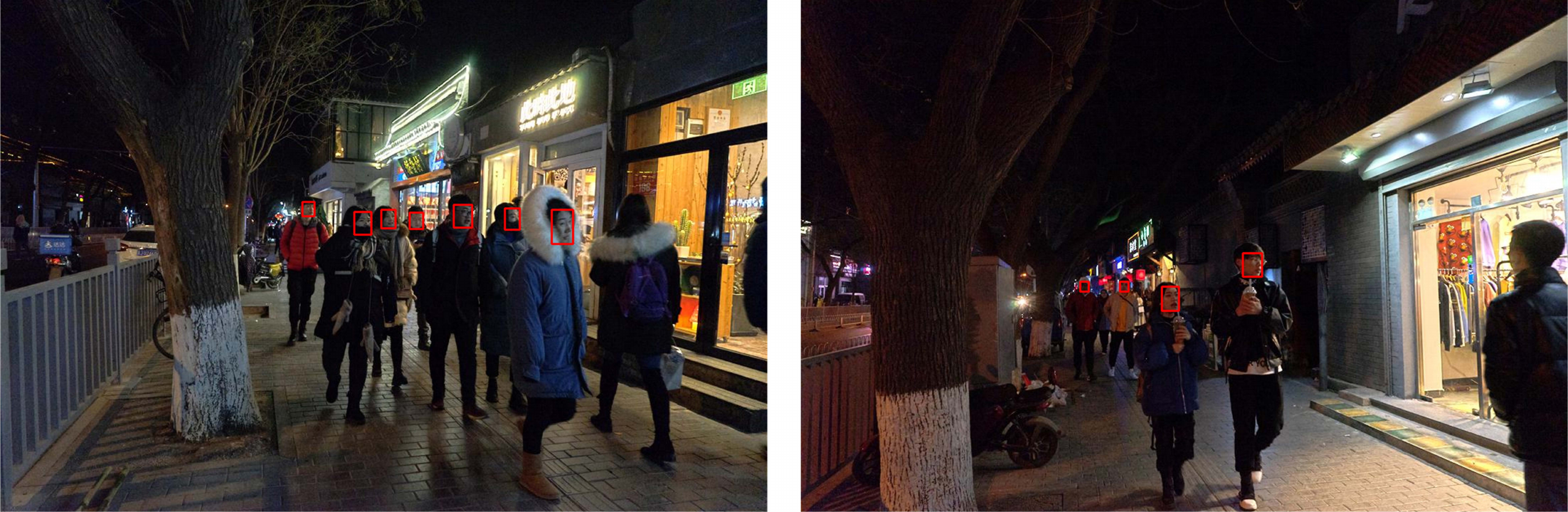} &
       \includegraphics[width=0.5\linewidth]{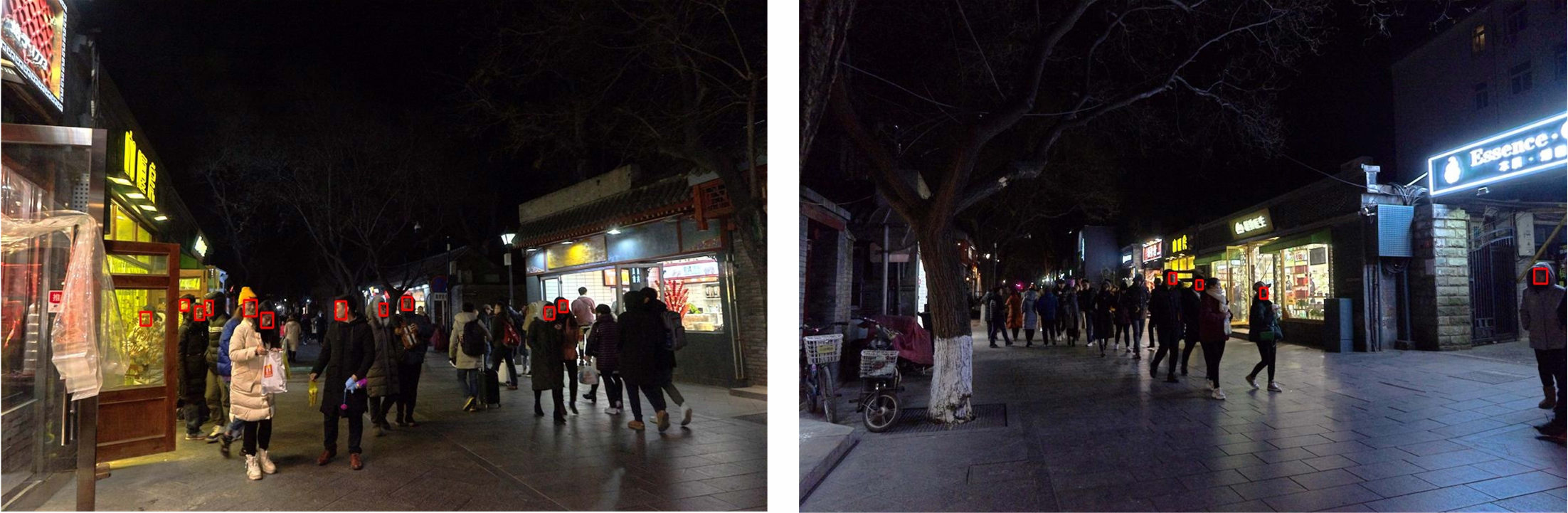} \\
        {\small(c)  RetinaFace\cite{deng2020retinaface} + ZERRINNet}    &
        {\small(d) DSFD\cite{li2019dsfd} +  ZERRINNet} \\
    \end{tabular}
    \caption{Extending our method to enhancement, the top row (low-light image) image is from darkface face detection dataset, where a,b are the results of low-light image detection from RetinaFace and DSFD face detection algorithms, respectively, and the bottom row (augmented image), where c,d are the results of low-light image augmentation from RetinaFace and DSFD face detection algorithms, respectively.}
    \label{face:image}
    \vspace{-3mm}
\end{figure*}

\begin{figure*}[!htbp]
\centering
\includegraphics[width=\textwidth]{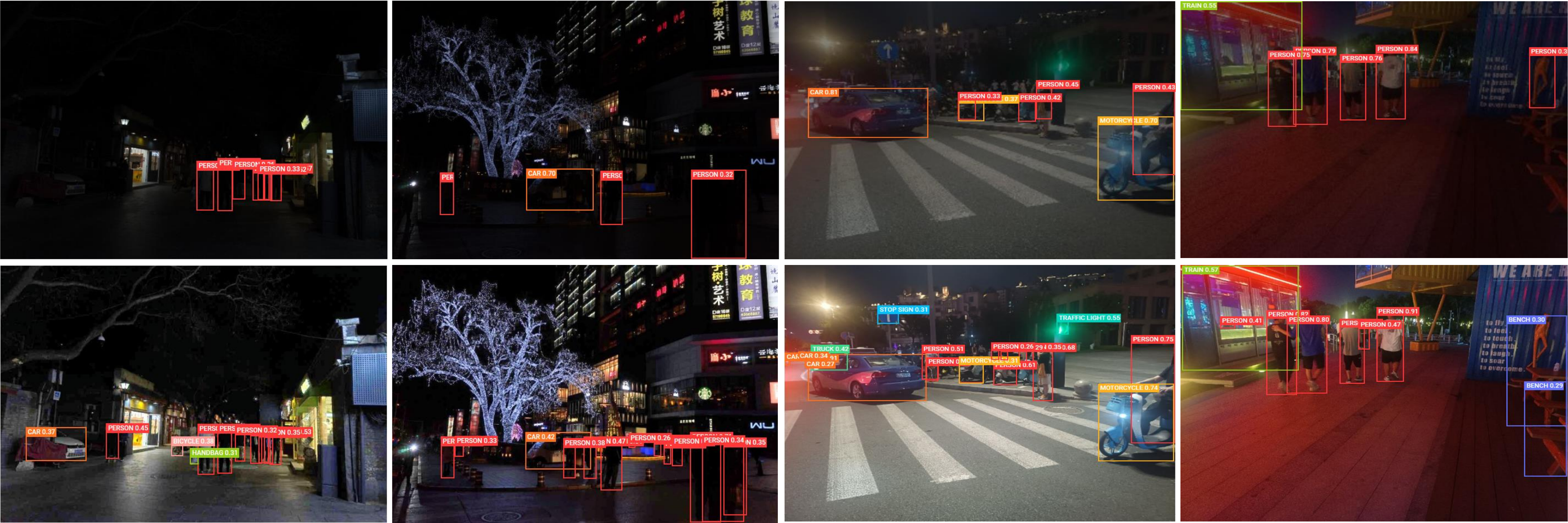}
\caption{Our method is applied to the target detection classification task, (top row) shows the low-light image detection results, (bottom row) shows the enhanced detection results, the two rows of images on the left are from the darkface public dataset, and the two rows on the right are real images taken at night scenes.}
\label{target_detection:image}
\end{figure*}

\begin{figure*}[!tbp]
\centering
\includegraphics[width=\textwidth]{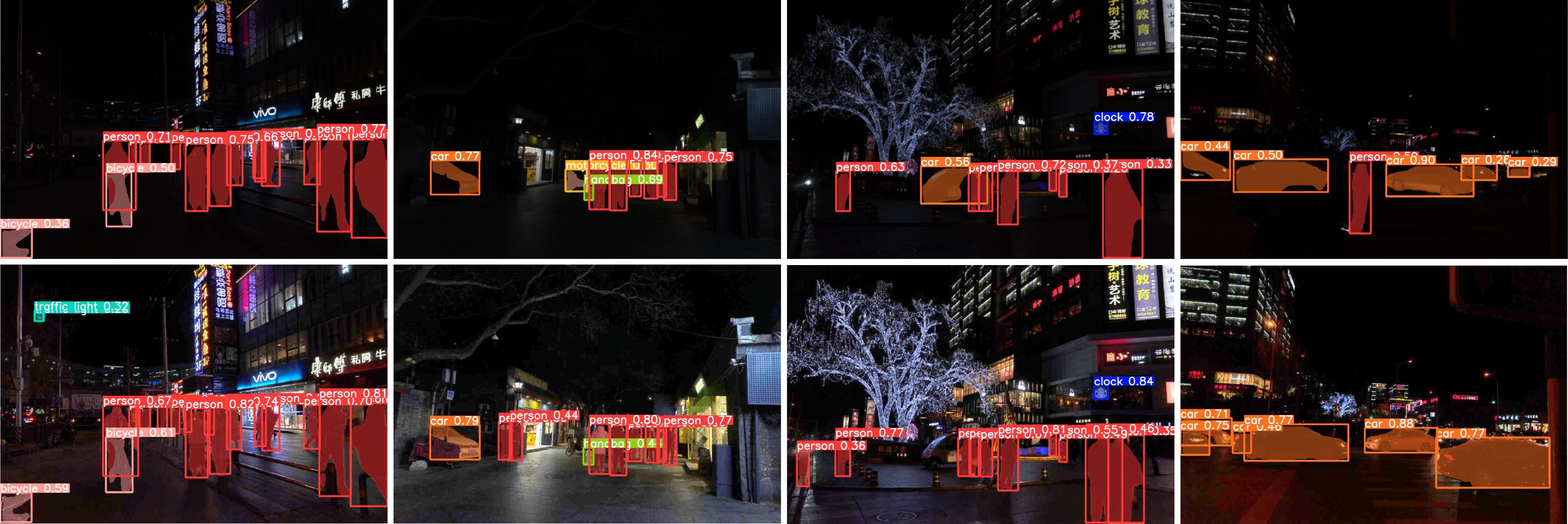}
\caption{Our method is applied to the instance segmentation task, (top row) shows the low-light image detection instance segmentation results, and (bottom row) shows the enhanced instance segmentation results.}
\label{instance_segmentation:image}
\end{figure*}

We produced a real shimmering dataset to test our method, our dataset was captured using a Huawei P30Pro cell phone, 25 shimmering images were captured at night, containing 5 grayscale images, 20 shimmering images. The 5 grayscale images are to validate the enhancement performance of the enhancement method on a single image, and the 20 color shimmering images which are rich in content and can be used for face detection, can be used for low-light image enhancement, target detection and instance segmentation for multiple tasks.
As shown in Fig.\ref{greyscale:figure} and Fig.\ref{colour:figure}, the enhancement results of low light gray scale low light images and real low light images are shown. The results show that the method has better enhancement results for monochrome low-light images and real low-light scenes, thus the method is stable and effective for general low-light image enhancement tasks.
Furthermore, we also apply our method to advanced vision tasks such as face detection, target detection, and instance segmentation, for the face detection task, we use RetinaFace\cite{deng2020retinaface} and DSFD\cite{li2019dsfd} face detection algorithms for the enhanced images, the dataset is from the darkface \cite{dackface} dataset, as shown in Fig. \ref{face:image}, the detection results show that our enhancement method is beneficial to help the face detection algorithm effect enhancement. In addition, for the task of target detection classification and instance segmentation, we use the current state-of-the-art detection algorithm yolov8x\cite{avazov2023yolov1-8} for the task, and the low-light images are from the darkface dataset and the nighttime low-light images of the real scene, respectively, and the results of the target detection classification and instance segmentation are shown in Figs \ref{target_detection:image} and \ref{instance_segmentation:image}, and the detection results show that without any pre-processing and fine-tuning, the detection results are very similar to the results of the face detection algorithm using our method, the performance of yolov8x is greatly improved. The above results show that our method shines on both low-level visual tasks and high-level visual tasks with good generalization performance.

\subsection{Analysis of The Impact of Loss Functions}
\begin{figure}[!tbp]
    \centering
    \tabcolsep=0.05cm
    \begin{tabular}{c c c }
        \includegraphics[width=0.33\linewidth]{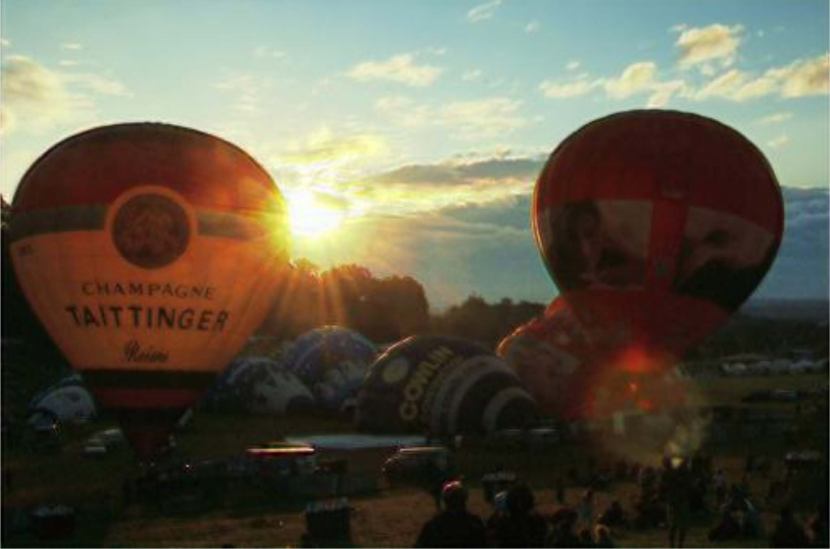} &
        \includegraphics[width=0.33\linewidth]{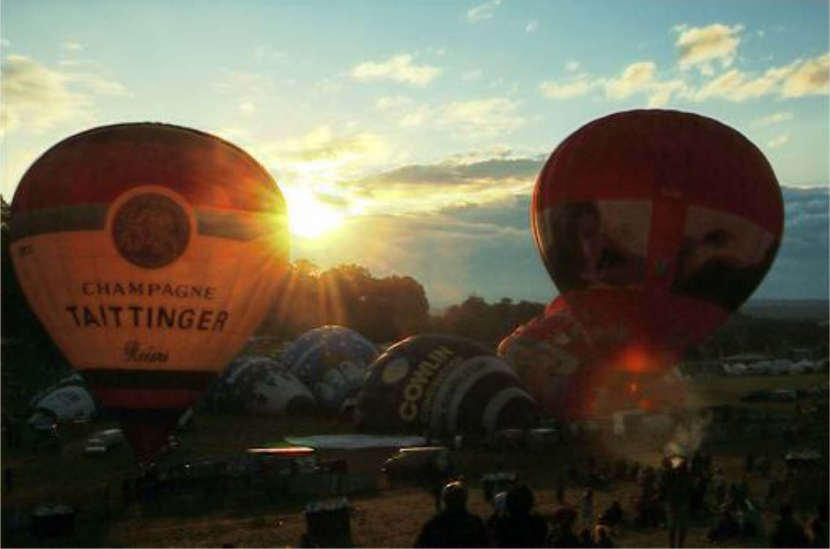} &
         \includegraphics[width=0.33\linewidth]{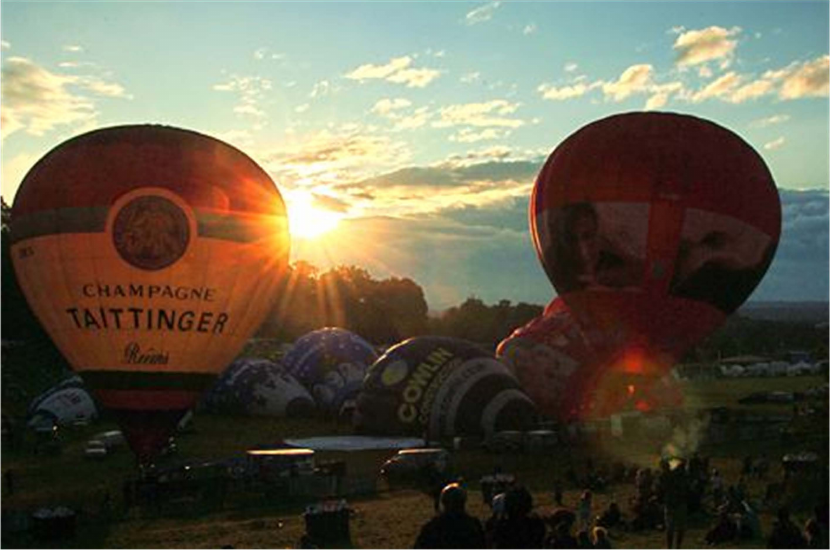} \\ 
          (a) & (b) & (c)  \\       
    \end{tabular}
    \caption{\small (a), (b), (c), represent the addition of the smoothing loss, the texture enhancement loss and the image noise loss, respectively, while the effect of the image satisfies the function of the loss function, which can be zoomed in to see the details.}
    \label{fig: loss1}
    \vspace{-3mm}
\end{figure}

In this section, the effects of network structure and loss function on the enhancement results are explored separately on the dataset NPE, as shown in Table \ref{tab:1}. Some visual comparisons are shown in Fig. \ref{fig: loss1}

\textit{1) loss functions:}
In order to study the impact of each loss function, we observe the change in performance by gradually adding the corresponding loss function, as shown in Table \ref{tab:1} (raw2-4), the default is only the reconstruction loss, and then add the smoothing loss, the image noise loss, and the texture enhancement loss in turn, it is clear that through the experiment we can find that all of the above losses, the performance enhancement are more helpful. As shown in Figure \ref{fig: loss1}, the niqe values of the corresponding images of the three are 4.3258, 3.6768 and 2.8943, respectively, and it can be noticed that the niqe values are gradually decreasing, the results show that the quality of the enhancement is improved by including more loss components.
\textit{2) Network structures:}  We evaluate the effectiveness of the above network tuning by appropriately adjusting the depth of the reflection network, as shown in Table \ref {tab:1} (row 5), taking into account that the reflection components hide more detailed information. Similarly to the loss function, the results show that we will bring better performance by increasing the depth of the reflection network appropriately.
\textit{3) Number of branches:} We explore the size of the number of convolutions corresponding to the decomposition and noise extraction networks
on the performance of image enhancement, as shown in Table \ref{tab:1}  (rows 6-7). Obviously, the increase of model size does not mean the improvement of image quality, in addition, increasing the model, will consume some computational resources, so we set the 6 branches to the default configuration.
\begin{table}[!tbp]
  \caption{Different settings and parameters of the network match the corresponding performance indicators}
  \centering
  \label{tab:1}
  \begin{tabular}{lccc}
    \toprule
    \textbf{Condition} & \textbf{NIQE} & \textbf{NIQMC}& \textbf{CPCQI} \\
    \midrule
    1. default configuration & 5.4400 & 3.2213 & 0.6089 \\
    \cmidrule{1-4} % 在第2列和第3列之间绘制横线
    2. with ${\mathcal{L}_{\mathrm{smooth}}}$ & 3.6721 & 5.2864 & 1.0177 \\
    3. with ${\mathcal{L}_{\mathrm{smooth}}}$+ with $\mathcal{L}_{\mathrm{texture}}$ & 3.4279 & 5.3339 & 1.0683 \\
    4. with $ \mathcal{L}_{\mathrm{smooth}}$+ with $\mathcal{L}_{\mathrm{texture}}$+with $ \mathcal{L}_{\mathrm{noise}}$ & 3.4109 & 5.3331 & 1.0701 \\
    \midrule % 在第2行之前添加双横线
    5. Reflection Network Depth & 3.4073 & 5.3393 & 1.0975 \\
    \midrule % 在第2行之后添加双横线
    6. Number of Convolutional Layers 5 & 3.4200 & 5.2989 & 1.0962 \\
    7. Number of Convolutional Layers 10 & 3.4311 & 5.2889 & 1.0846 \\
    \bottomrule
  \end{tabular}
\end{table}
\section{CONCLUSION}
In this paper, we present a learning-based Retinex zero-sample enhancement method that simultaneously addresses the problems of low-light image contrast, noise, color distortion, and poor generalization ability of existing enhancement methods. The proposed method outperforms competitive algorithms on seven challenging datasets without requiring any external datasets for training.  Moreover, we conduct experiments on real datasets and face detection datasets produced by ourselves, and the results show that the method exhibits better generalization performance in gray map enhancement, real shimmering images, face detection, target recognition, and instance segmentation. In the future, we will further optimize the enhancement speed of our method and move from shimmer image enhancement to shimmer video enhancement.
%\section*{Acknowledgments}
%This should be a simple paragraph before the References to thank those individuals and institutions who have supported your work on this article.

%{\appendix[Proof of the Zonklar Equations]
%Use $\backslash${\tt{appendix}} if you have a single appendix:
%Do not use $\backslash${\tt{section}} anymore after $\backslash${\tt{appendix}}, only $\backslash${\tt{section*}}.
%If you have multiple appendixes use $\backslash${\tt{appendices}} then use $\backslash${\tt{section}} to start each appendix.
%You must declare a $\backslash${\tt{section}} before using any $\backslash${\tt{subsection}} or using $\backslash${\tt{label}} ($\backslash${\tt{appendices}} by itself
 %starts a section numbered zero.)}


%\section{References Section}
\bibliographystyle{IEEEtran}
\bibliography{zerrinnet}

\begin{thebibliography}{10}
\providecommand{\url}[1]{#1}
\csname url@samestyle\endcsname
\providecommand{\newblock}{\relax}
\providecommand{\bibinfo}[2]{#2}
\providecommand{\BIBentrySTDinterwordspacing}{\spaceskip=0pt\relax}
\providecommand{\BIBentryALTinterwordstretchfactor}{4}
\providecommand{\BIBentryALTinterwordspacing}{\spaceskip=\fontdimen2\font plus
\BIBentryALTinterwordstretchfactor\fontdimen3\font minus
  \fontdimen4\font\relax}
\providecommand{\BIBforeignlanguage}[2]{{%
\expandafter\ifx\csname l@#1\endcsname\relax
\typeout{** WARNING: IEEEtran.bst: No hyphenation pattern has been}%
\typeout{** loaded for the language `#1'. Using the pattern for}%
\typeout{** the default language instead.}%
\else
\language=\csname l@#1\endcsname
\fi
#2}}
\providecommand{\BIBdecl}{\relax}
\BIBdecl

\bibitem{zheng2022low}
S.~Zheng, Y.~Ma, J.~Pan, C.~Lu, and G.~Gupta, ``Low-light image and video
  enhancement: A comprehensive survey and beyond,'' \emph{arXiv preprint
  arXiv:2212.10772}, 2022.

\bibitem{2021survry}
C.~Li, C.~Guo, L.~Han, J.~Jiang, M.-M. Cheng, J.~Gu, and C.~C. Loy,
  ``Low\-light image and video enhancement using deep learning: A survey,''
  \emph{IEEE transactions on pattern analysis and machine intelligence},
  vol.~44, no.~12, pp. 9396--9416, 2021.

\bibitem{land1971lightness}
E.~H. Land and J.~J. McCann, ``Lightness and retinex theory,'' \emph{Josa},
  vol.~61, no.~1, pp. 1\--11, 1971.

\bibitem{MSR}
Z.-u. Rahman, D.~J. Jobson, and G.~A. Woodell, ``Multi-scale retinex for color
  image enhancement,'' in \emph{Proceedings of 3rd IEEE international
  conference on image processing}, vol.~3.\hskip 1em plus 0.5em minus
  0.4em\relax IEEE, 1996, pp. 1003--1006.

\bibitem{MSRCR}
D.~J. Jobson, Z.-u. Rahman, and G.~A. Woodell, ``A multiscale retinex for
  bridging the gap between color images and the human observation of scenes,''
  \emph{IEEE Transactions on Image processing}, vol.~6, no.~7, pp. 965--976,
  1997.

\bibitem{zhang2021better}
Y.~Zhang, X.~Di, B.~Zhang, R.~Ji, and C.~Wang, ``Better than reference in
  low-light image enhancement: conditional re-enhancement network,'' \emph{IEEE
  Transactions on Image Processing}, vol.~31, pp. 759--772, 2021.

\bibitem{retinex-net}
C.~Wei, W.~Wang, W.~Yang, and J.~Liu, ``Deep retinex decomposition for
  low-light enhancement,'' \emph{arXiv preprint arXiv:1808.04560}, 2018.

\bibitem{uretinex}
W.~Wu, J.~Weng, P.~Zhang, X.~Wang, W.~Yang, and J.~Jiang, ``Uretinex-net:
  Retinex-based deep unfolding network for low-light image enhancement,'' in
  \emph{Proceedings of the IEEE/CVF conference on computer vision and pattern
  recognition}, 2022, pp. 5901--5910.

\bibitem{reza2004realization}
A.~M. Reza, ``Realization of the contrast limited adaptive histogram
  equalization (clahe) for real\-time image enhancement,'' \emph{Journal of
  VLSI signal processing systems for signal, image and video technology},
  vol.~38, pp. 35\--44, 2004.

\bibitem{hedenoiseimage}
R.~Dorothy, R.~Joany, R.~J. Rathish, S.~S. Prabha, S.~Rajendran, and S.~Joseph,
  ``Image enhancement by histogram equalization,'' \emph{International Journal
  of Nano Corrosion Science and Engineering}, vol.~2, no.~4, pp. 21\--30, 2015.

\bibitem{hebrightness}
C.~Wang and Z.~Ye, ``Brightness preserving histogram equalization with maximum
  entropy: a variational perspective,'' \emph{IEEE Transactions on Consumer
  Electronics}, vol.~51, no.~4, pp. 1326\--1334, 2005.

\bibitem{dackhe}
S.-C. Huang, F.-C. Cheng, and Y.-S. Chiu, ``Efficient contrast enhancement
  using adaptive gamma correction with weighting distribution,'' \emph{IEEE
  transactions on image processing}, vol.~22, no.~3, pp. 1032\--1041, 2012.

\bibitem{pizer1987adaptive}
S.~M. Pizer, E.~P. Amburn, J.~D. Austin, R.~Cromartie, A.~Geselowitz, T.~Greer,
  B.~ter Haar~Romeny, J.~B. Zimmerman, and K.~Zuiderveld, ``Adaptive histogram
  equalization and its variations,'' \emph{Computer vision, graphics, and image
  processing}, vol.~39, no.~3, pp. 355\--368, 1987.

\bibitem{wang2013}
S.~Wang, J.~Zheng, H.-M. Hu, and B.~Li, ``Naturalness preserved enhancement
  algorithm for non\-uniform illumination images,'' \emph{IEEE transactions on
  image processing}, vol.~22, no.~9, pp. 3538\--3548, 2013.

\bibitem{LIME}
X.~Guo, Y.~Li, and H.~Ling, ``Lime: Low\-light image enhancement via
  illumination map estimation,'' \emph{IEEE Transactions on image processing},
  vol.~26, no.~2, pp. 982\--993, 2016.

\bibitem{Mading}
M.~Li, J.~Liu, W.~Yang, X.~Sun, and Z.~Guo, ``Structure\-revealing low-light
  image enhancement via robust retinex model,'' \emph{IEEE Transactions on
  Image Processing}, vol.~27, no.~6, pp. 2828\--2841, 2018.

\bibitem{li2021low}
J.~Li, X.~Feng, and Z.~Hua, ``Low-light image enhancement via
  progressive-recursive network,'' \emph{IEEE Transactions on Circuits and
  Systems for Video Technology}, vol.~31, no.~11, pp. 4227--4240, 2021.

\bibitem{ren2018lecarm}
Y.~Ren, Z.~Ying, T.~H. Li, and G.~Li, ``Lecarm: Low-light image enhancement
  using the camera response model,'' \emph{IEEE Transactions on Circuits and
  Systems for Video Technology}, vol.~29, no.~4, pp. 968--981, 2018.

\bibitem{LLNet}
K.~G. Lore, A.~Akintayo, and S.~Sarkar, ``Llnet: A deep autoencoder approach to
  natural low-light image enhancement,'' \emph{Pattern Recognition}, vol.~61,
  pp. 650\--662, 2017.

\bibitem{MBLLEN}
F.~Lv, F.~Lu, J.~Wu, and C.~Lim, ``Mbllen: Low\-light image/video enhancement
  using cnns.'' in \emph{BMVC}, vol. 220, no.~1, 2018, p.~4.

\bibitem{attention}
F.~Lv, Y.~Li, and F.~Lu, ``Attention guided low\-light image enhancement with a
  large scale low-light simulation dataset,'' \emph{International Journal of
  Computer Vision}, vol. 129, no.~7, pp. 2175\--2193, 2021.

\bibitem{xu2020learning}
K.~Xu, X.~Yang, B.~Yin, and R.~W. Lau, ``Learning to restore low\-light images
  via decomposition-and-enhancement,'' in \emph{Proceedings of the IEEE/CVF
  conference on computer vision and pattern recognition}, 2020, pp.
  2281\--2290.

\bibitem{utvnet}
C.~Zheng, D.~Shi, and W.~Shi, ``Adaptive unfolding total variation network for
  low\-light image enhancement,'' in \emph{Proceedings of the IEEE/CVF
  international conference on computer vision}, 2021, pp. 4439\--4448.

\bibitem{llFlow}
Y.~Wang, R.~Wan, W.~Yang, H.~Li, L.-P. Chau, and A.~Kot, ``Low\-light image
  enhancement with normalizing flow,'' in \emph{Proceedings of the AAAI
  conference on artificial intelligence}, vol.~36, no.~3, 2022, pp.
  2604\--2612.

\bibitem{DRBN}
W.~Yang, S.~Wang, Y.~Fang, Y.~Wang, and J.~Liu, ``From fidelity to perceptual
  quality: A semi-supervised approach for low\-light image enhancement,'' in
  \emph{Proceedings of the IEEE/CVF conference on computer vision and pattern
  recognition}, 2020, pp. 3063\--3072.

\bibitem{enlightengan}
Y.~Jiang, X.~Gong, D.~Liu, Y.~Cheng, C.~Fang, X.~Shen, J.~Yang, P.~Zhou, and
  Z.~Wang, ``Enlightengan: Deep light enhancement without paired supervision,''
  \emph{IEEE transactions on image processing}, vol.~30, pp. 2340\--2349, 2021.

\bibitem{sci}
L.~Ma, T.~Ma, R.~Liu, X.~Fan, and Z.~Luo, ``Toward fast, flexible, and robust
  low-light image enhancement,'' in \emph{Proceedings of the IEEE/CVF
  Conference on Computer Vision and Pattern Recognition}, 2022, pp.
  5637\--5646.

\bibitem{yu2017low}
S.-Y. Yu and H.~Zhu, ``Low-illumination image enhancement algorithm based on a
  physical lighting model,'' \emph{IEEE Transactions on Circuits and Systems
  for Video Technology}, vol.~29, no.~1, pp. 28--37, 2017.

\bibitem{xu2022structure}
K.~Xu, H.~Chen, C.~Xu, Y.~Jin, and C.~Zhu, ``Structure-texture aware network
  for low-light image enhancement,'' \emph{IEEE Transactions on Circuits and
  Systems for Video Technology}, vol.~32, no.~8, pp. 4983--4996, 2022.

\bibitem{rrdnet}
A.~Zhu, L.~Zhang, Y.~Shen, Y.~Ma, S.~Zhao, and Y.~Zhou, ``Zero\-shot
  restoration of underexposed images via robust retinex decomposition,'' in
  \emph{2020 IEEE International Conference on Multimedia and Expo
  (ICME)}.\hskip 1em plus 0.5em minus 0.4em\relax IEEE, 2020, pp. 1\--6.

\bibitem{liang2022self}
J.~Liang, Y.~Xu, Y.~Quan, B.~Shi, and H.~Ji, ``Self-supervised low-light image
  enhancement using discrepant untrained network priors,'' \emph{IEEE
  Transactions on Circuits and Systems for Video Technology}, vol.~32, no.~11,
  pp. 7332--7345, 2022.

\bibitem{LightenNet}
C.~Li, J.~Guo, F.~Porikli, and Y.~Pang, ``Lightennet: A convolutional neural
  network for weakly illuminated image enhancement,'' \emph{Pattern recognition
  letters}, vol. 104, pp. 15\--22, 2018.

\bibitem{kind}
Y.~Zhang, J.~Zhang, and X.~Guo, ``Kindling the darkness: A practical low-light
  image enhancer,'' in \emph{Proceedings of the 27th ACM international
  conference on multimedia}, 2019, pp. 1632\--1640.

\bibitem{kind++}
Y.~Zhang, X.~Guo, J.~Ma, W.~Liu, and J.~Zhang, ``Beyond brightening low-light
  images,'' \emph{International Journal of Computer Vision}, vol. 129, pp.
  1013\--1037, 2021.

\bibitem{wang}
Y.~Wang, Y.~Cao, Z.-J. Zha, J.~Zhang, Z.~Xiong, W.~Zhang, and F.~Wu,
  ``Progressive retinex: Mutually reinforced illumination-noise perception
  network for low-light image enhancement,'' in \emph{Proceedings of the 27th
  ACM international conference on multimedia}, 2019, pp. 2015--2023.

\bibitem{fan_segmentation}
M.~Fan, W.~Wang, W.~Yang, and J.~Liu, ``Integrating semantic segmentation and
  retinex model for low-light image enhancement,'' in \emph{Proceedings of the
  28th ACM international conference on multimedia}, 2020, pp. 2317--2325.

\bibitem{ExCNet}
L.~Zhang, L.~Zhang, X.~Liu, Y.~Shen, S.~Zhang, and S.~Zhao, ``Zero-shot
  restoration of back-lit images using deep internal learning,'' in
  \emph{Proceedings of the 27th ACM international conference on multimedia},
  2019, pp. 1623--1631.

\bibitem{guo2020zero}
C.~Guo, C.~Li, J.~Guo, C.~C. Loy, J.~Hou, S.~Kwong, and R.~Cong,
  ``Zero\-reference deep curve estimation for low-light image enhancement,'' in
  \emph{Proceedings of the IEEE/CVF conference on computer vision and pattern
  recognition}, 2020, pp. 1780\--1789.

\bibitem{retinexdip}
Z.~Zhao, B.~Xiong, L.~Wang, Q.~Ou, L.~Yu, and F.~Kuang, ``Retinexdip: A unified
  deep framework for low-light image enhancement,'' \emph{IEEE Transactions on
  Circuits and Systems for Video Technology}, vol.~32, no.~3, pp. 1076--1088,
  2021.

\bibitem{Zero-DCE++}
C.~Li, C.~Guo, and C.~C. Loy, ``Learning to enhance low-light image via
  zero-reference deep curve estimation,'' \emph{IEEE Transactions on Pattern
  Analysis and Machine Intelligence}, vol.~44, no.~8, pp. 4225--4238, 2021.

\bibitem{RUAS}
R.~Liu, L.~Ma, J.~Zhang, X.~Fan, and Z.~Luo, ``Retinex-inspired unrolling with
  cooperative prior architecture search for low-light image enhancement,'' in
  \emph{Proceedings of the IEEE/CVF Conference on Computer Vision and Pattern
  Recognition}, 2021, pp. 10\,561--10\,570.

\bibitem{SGZ}
S.~Zheng and G.~Gupta, ``Semantic-guided zero-shot learning for low-light
  image/video enhancement,'' in \emph{Proceedings of the IEEE/CVF Winter
  conference on applications of computer vision}, 2022, pp. 581--590.

\bibitem{hai2023r2rnet}
J.~Hai, Z.~Xuan, R.~Yang, Y.~Hao, F.~Zou, F.~Lin, and S.~Han, ``R2rnet:
  Low\-light image enhancement via real-low to real-normal network,''
  \emph{Journal of Visual Communication and Image Representation}, vol.~90, p.
  103712, 2023.

\bibitem{xu2012structure}
L.~Xu, Q.~Yan, Y.~Xia, and J.~Jia, ``Structure extraction from texture via
  relative total variation,'' \emph{ACM transactions on graphics (TOG)},
  vol.~31, no.~6, pp. 1\--10, 2012.

\bibitem{wang2019underexposed}
R.~Wang, Q.~Zhang, C.-W. Fu, X.~Shen, W.-S. Zheng, and J.~Jia, ``Underexposed
  photo enhancement using deep illumination estimation,'' in \emph{Proceedings
  of the IEEE/CVF conference on computer vision and pattern recognition}, 2019,
  pp. 6849\--6857.

\bibitem{6467022}
C.~Lee, C.~Lee, and C.-S. Kim, ``Contrast enhancement based on layered
  difference representation,'' in \emph{2012 19th IEEE International Conference
  on Image Processing}, 2012, pp. 965--968.

\bibitem{loh2019getting}
Y.~P. Loh and C.~S. Chan, ``Getting to know low-light images with the
  exclusively dark dataset,'' \emph{Computer Vision and Image Understanding},
  vol. 178, pp. 30--42, 2019.

\bibitem{fu2016fusion}
X.~Fu, D.~Zeng, Y.~Huang, Y.~Liao, X.~Ding, and J.~Paisley, ``A fusion\-based
  enhancing method for weakly illuminated images,'' \emph{Signal Processing},
  vol. 129, pp. 82\--96, 2016.

\bibitem{7782813}
X.~Guo, Y.~Li, and H.~Ling, ``Lime: Low\-light image enhancement via
  illumination map estimation,'' \emph{IEEE Transactions on Image Processing},
  vol.~26, no.~2, pp. 982--993, 2017.

\bibitem{7120119}
K.~Ma, K.~Zeng, and Z.~Wang, ``Perceptual quality assessment for multi-exposure
  image fusion,'' \emph{IEEE Transactions on Image Processing}, vol.~24,
  no.~11, pp. 3345--3356, 2015.

\bibitem{6512558}
S.~Wang, J.~Zheng, H.-M. Hu, and B.~Li, ``Naturalness preserved enhancement
  algorithm for non-uniform illumination images,'' \emph{IEEE Transactions on
  Image Processing}, vol.~22, no.~9, pp. 3538--3548, 2013.

\bibitem{1}
A.~Mittal, R.~Soundararajan, and A.~C. Bovik, ``Making a “completely blind”
  image quality analyzer,'' \emph{IEEE Signal processing letters}, vol.~20,
  no.~3, pp. 209--212, 2012.

\bibitem{2}
K.~Gu, D.~Tao, J.-F. Qiao, and W.~Lin, ``Learning a no\-reference quality
  assessment model of enhanced images with big data,'' \emph{IEEE transactions
  on neural networks and learning systems}, vol.~29, no.~4, pp. 1301--1313,
  2017.

\bibitem{gu2016no}
K.~Gu, W.~Lin, G.~Zhai, X.~Yang, W.~Zhang, and C.~W. Chen, ``No\-reference
  quality metric of contrast-distorted images based on information
  maximization,'' \emph{IEEE transactions on cybernetics}, vol.~47, no.~12, pp.
  4559--4565, 2016.

\bibitem{Zero-DCE}
L.~Zhang, L.~Zhang, X.~Liu, Y.~Shen, S.~Zhang, and S.~Zhao, ``Zero-shot
  restoration of back-lit images using deep internal learning,'' in
  \emph{Proceedings of the 27th ACM international conference on multimedia},
  2019, pp. 1623--1631.

\bibitem{deng2020retinaface}
J.~Deng, J.~Guo, E.~Ververas, I.~Kotsia, and S.~Zafeiriou, ``Retinaface:
  Single\-shot multi\-level face localisation in the wild,'' in
  \emph{Proceedings of the IEEE/CVF conference on computer vision and pattern
  recognition}, 2020, pp. 5203\--5212.

\bibitem{li2019dsfd}
J.~Li, Y.~Wang, C.~Wang, Y.~Tai, J.~Qian, J.~Yang, C.~Wang, J.~Li, and
  F.~Huang, ``Dsfd: dual shot face detector,'' in \emph{Proceedings of the
  IEEE/CVF Conference on Computer Vision and Pattern Recognition}, 2019, pp.
  5060\--5069.

\bibitem{dackface}
Y.~Yuan, W.~Yang, W.~Ren, J.~Liu, W.~J. Scheirer, and Z.~Wang, ``Ug
  $\hat{}$\{$2+$\}$ $ track 2: A collective benchmark effort for evaluating and
  advancing image understanding in poor visibility environments,'' \emph{arXiv
  preprint arXiv:1904.04474}, 2019.

\bibitem{avazov2023yolov1-8}
K.~Avazov, M.~K. Jamil, B.~Muminov, A.~B. Abdusalomov, and Y.-I. Cho, ``Fire
  detection and notification method in ship areas using deep learning and
  computer vision approaches,'' \emph{Sensors}, vol.~23, no.~16, p. 7078, 2023.

\end{thebibliography}
%\begin{thebibliography}{1}
%\end{thebibliography}

%\newpage

%\bf{If you include a photo:}\vspace{-33pt}
\begin{IEEEbiography}[{\includegraphics[width=1in,height=1.25in,clip,keepaspectratio]{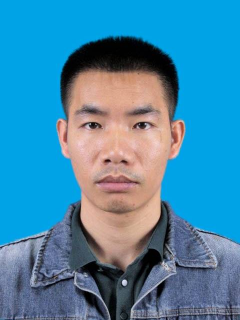}}]{WenChao Li}
was born in Jiangxi. He received the B.S. and M.S. degrees from Nanchang University of Aeronautics and Astronautics, China, in 2018 and 2021, respectively, and is currently pursuing the Ph.D. degree at Beijing University of Aeronautics and Astronautics. His research interests include image restoration and enhancement and helicopter dynamic parameter measurement.
\end{IEEEbiography}

\begin{IEEEbiography}[{\includegraphics[width=1in,height=1.25in,clip,keepaspectratio]{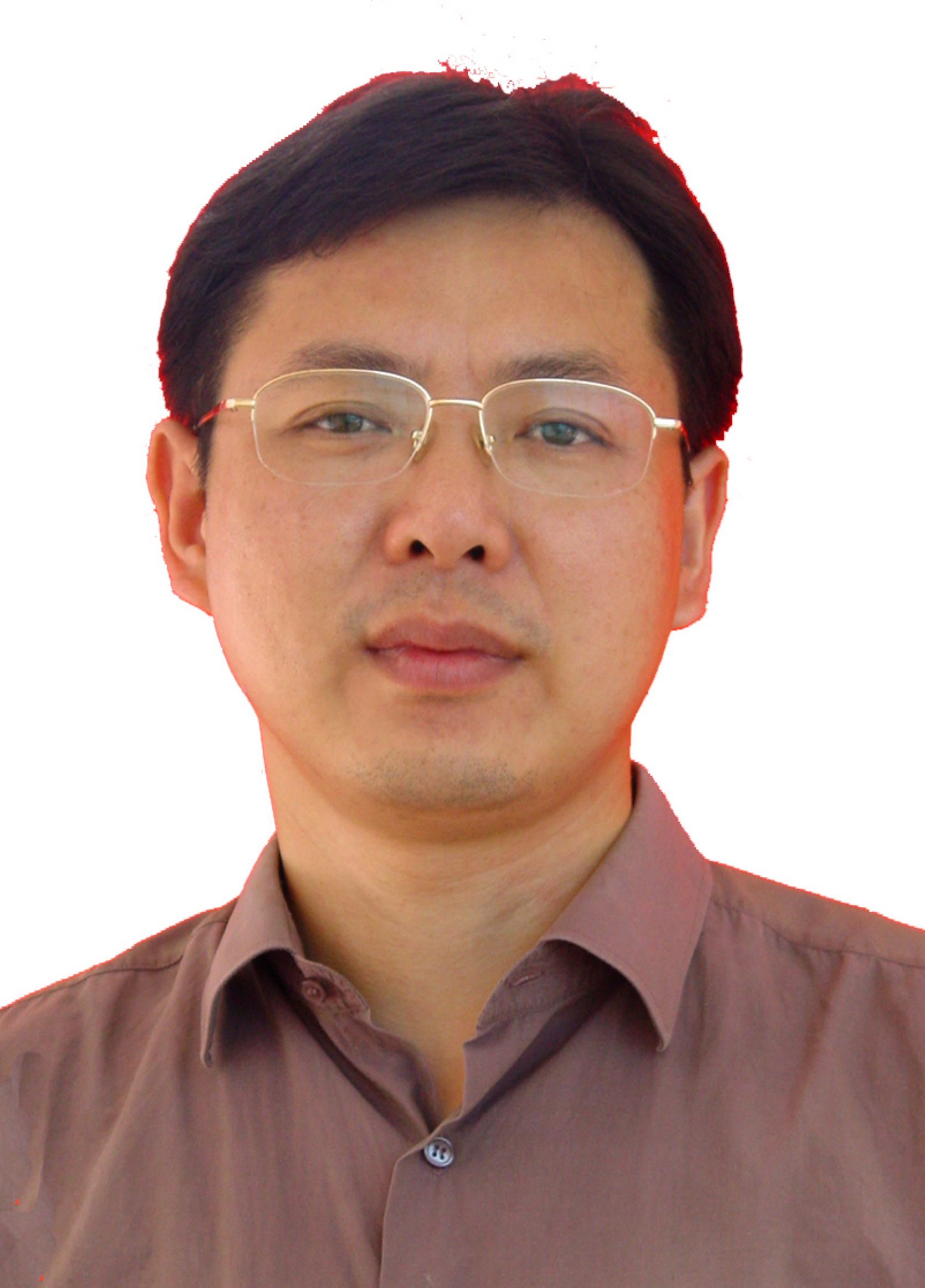}}]{Bangshu Xiong}
received the M.S. and Ph.D. degrees in signal and information processing from North-western Polytechnical University, Xi’an, in 2001 and 2004, respectively. He is currently a Professor with the School of Information Engineering, Nanchang Hangkong University, China. His research interests include image processing, computer vision, and helicopter fault diagnosis technology. He was a recipient of the Natural Science Award (Second class) of Jiangxi Province of China, and the Scientific and Technological Achievement Award (Second class) in higher institution of Jiangxi Province of China.
\end{IEEEbiography}

\begin{IEEEbiography}[{\includegraphics[width=1in,height=1.25in,clip,keepaspectratio]{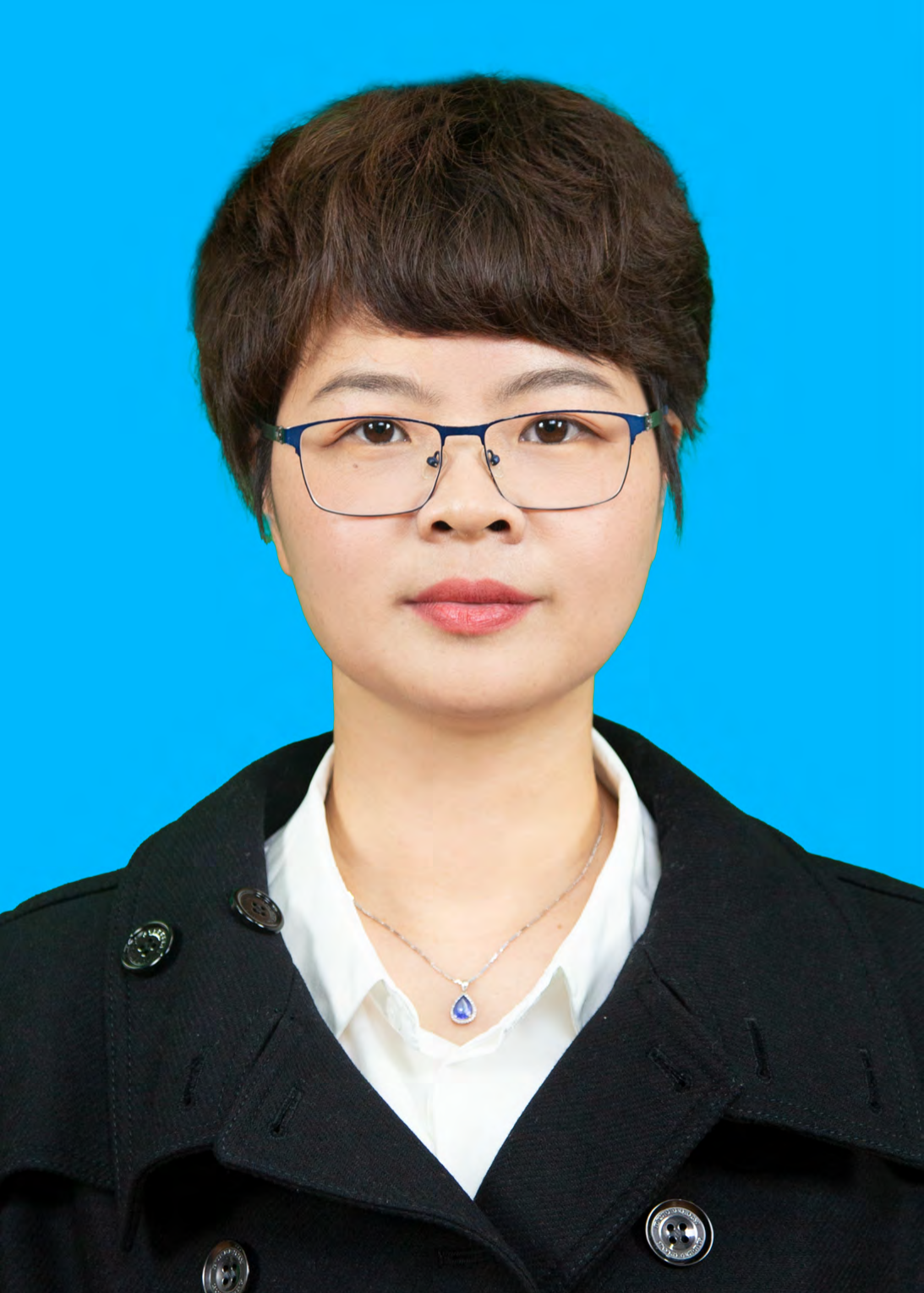}}]{Qiaofeng Ou}
received the B.S. degree in communication engineer-ing, and the M.S. and Ph.D. degrees in information and communication engineering from Northwestern Polytechnical University, Xi’an, China, in 2004, 2007, and 2018, respectively. In 2007, she joined as a Lecturer with Nanchang Hang-kong University. Her research interest includes image processing.
\end{IEEEbiography}

\begin{IEEEbiography}[{\includegraphics[width=1in,height=1.25in,clip,keepaspectratio]{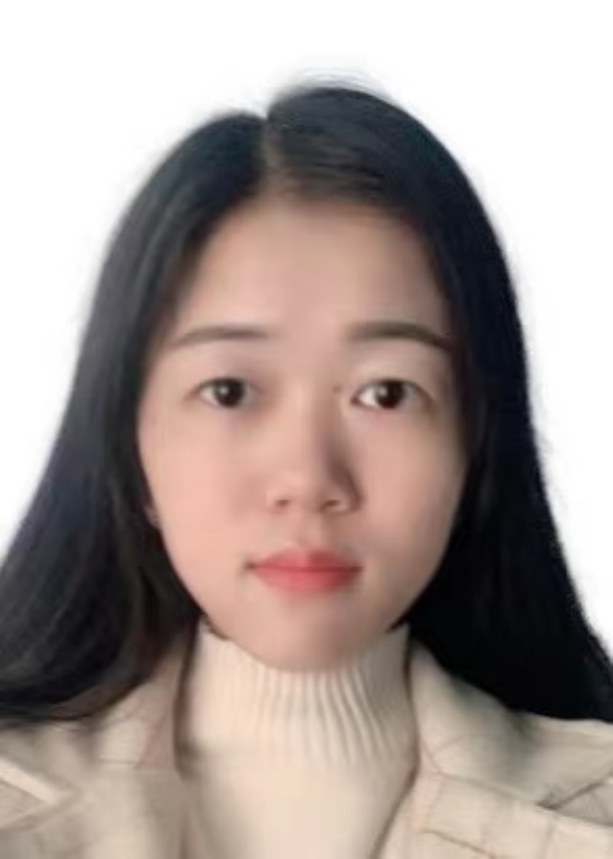}}]{Xiaoyun Long}
 was born in Jiangxi , China. She received the B.E. degree in Electronic Information Engineering from Nanchang Aviation University in 2018 and the M.S. degree in Electronic and Communication Engineering  from Nanchang University in 2022. She is currently working at Ganzhou Vocational and Technical College.Her research interest includes Radar signal processing,image restoration, and image enhancement.
\end{IEEEbiography}

\begin{IEEEbiography}[{\includegraphics[width=1in,height=1.25in,clip,keepaspectratio]{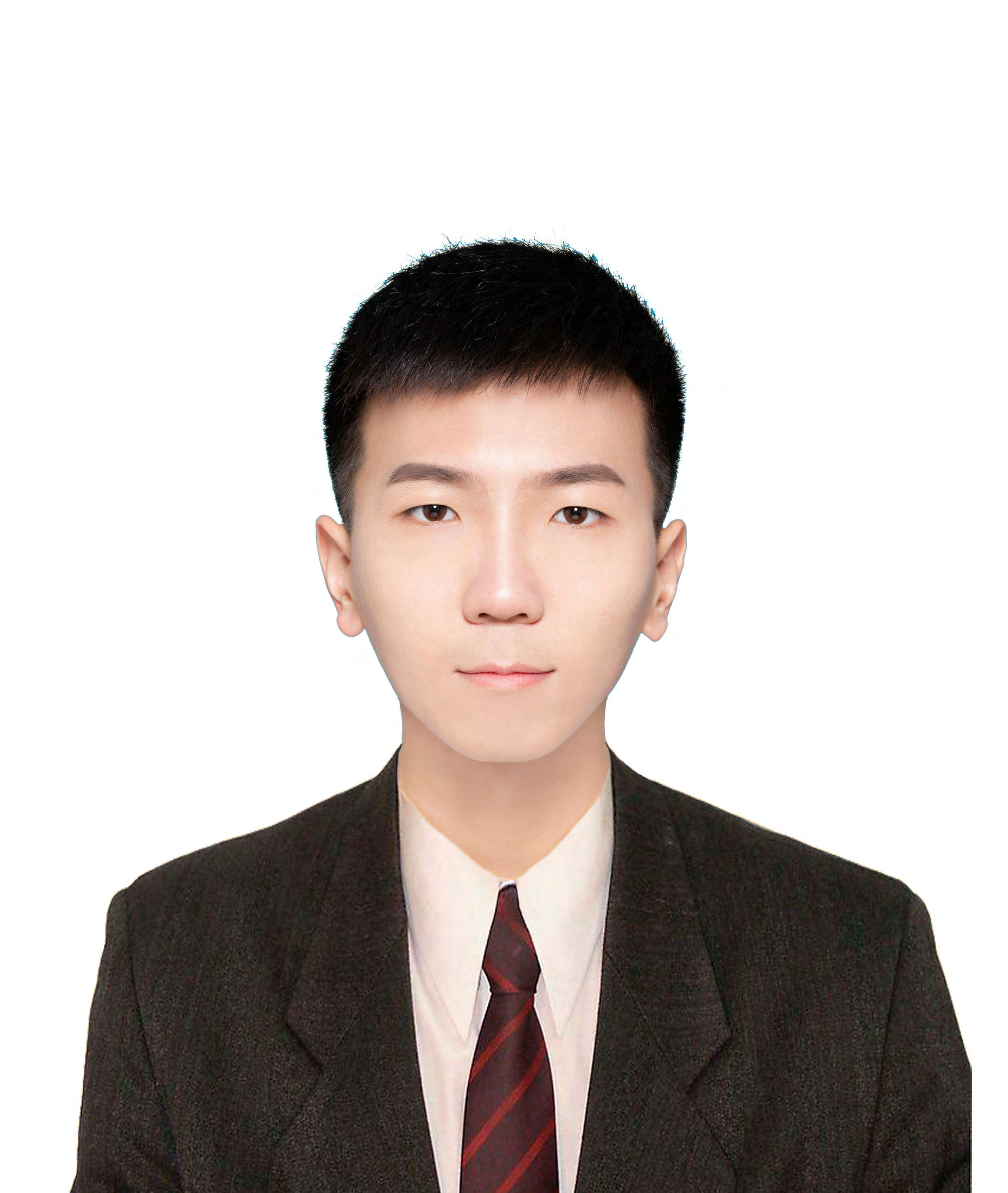}}]{Jinhao Zhu}
was born in Hebei, China. He received the B.E. degree in Electronics Science and Technology from Hebei University of Technology in 2020 and an M.S. degree in Astronomy from the University of Chinese Academy of Sciences in 2023. He is currently studying for a Ph.D degree in Instrument Science and Technology in BeiHang University. His research interests include image enhancement and recognition.
\end{IEEEbiography}

\begin{IEEEbiography}[{\includegraphics[width=1in,height=1.25in,clip,keepaspectratio]{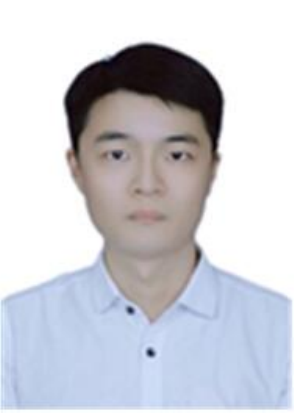}}]{Shuyuan Wen}
was born in Shandong, China, in 1998. He received the B.S. degree from School of Control Science and Engineering, China University of Petroleum (East China), Qingdao, China, in 2021. He is currently pursuing the Ph.D. degree in the School of Instrumentation Science and Opto-Electronics Engineering and the SHENYUAN Honors College, Beihang University, Beijing, China. His current research interests include computer Vision, pattern recognition and vision measurement.
\end{IEEEbiography}

\begin{IEEEbiography}[{\includegraphics[width=1in,height=1.25in,clip,keepaspectratio]{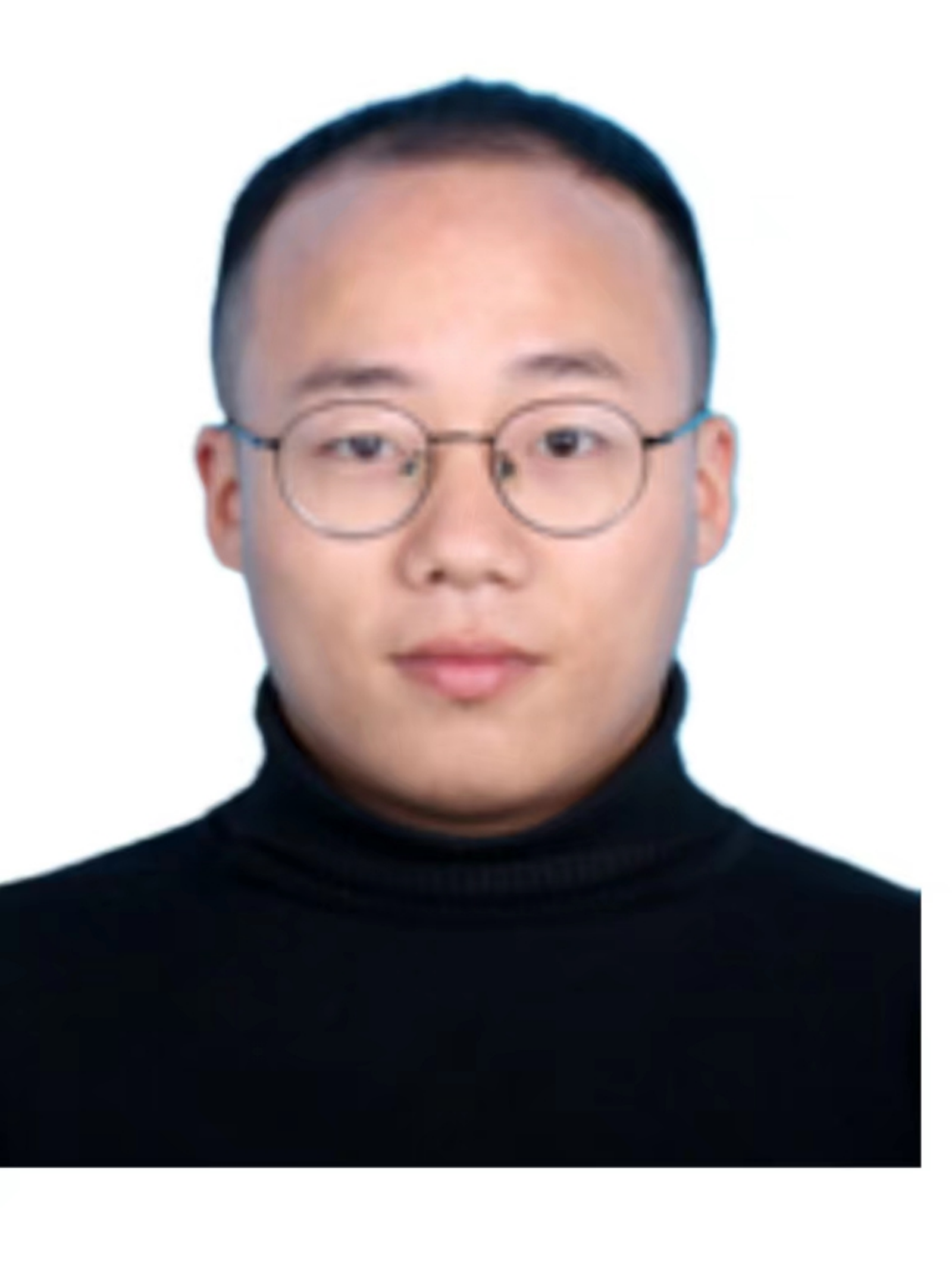}}]
{jiabao Chen}
was born in Jiujiang, Jiangxi, China, in 1998. He received the B.Eng. degree from Nanchang Hangkong University, where he is currently pursuing the M.S. degree with the School of Information Engineering. His research interests include machine learning, object detection.
\end{IEEEbiography}

%\vfill

\end{document}